\pdfoutput=1 
\documentclass[subscriptcorrection,upint,varvw, nolists]{asmejour}
\usepackage{multirow}
\usepackage{makecell}

\hypersetup{%
	pdfauthor={Enrico Halim, Hemant Kumar Singh, Tapabrata Ray},
	pdftitle={CR-EA-C: A Confidence-Driven Evolutionary Algorithm for Noisy Optimization with Joint Chance Constraints},
	pdfkeywords={chance constrained optimization, noisy optimization, evolutionary algorithms, stochastic programming},
	pdfsubject={Optimization under uncertainty using evolutionary algorithms},
}

\begin{document}


\SetAuthorBlock{Enrico Halim\CorrespondingAuthor}{%
School of Engineering and Technology,\\
The University of New South Wales,\\
Canberra, Australia \\
email: enrico.halim@unsw.edu.au
}

\SetAuthorBlock{Hemant Kumar Singh}{%
School of Engineering and Technology,\\
The University of New South Wales,\\
Canberra, Australia \\
email: hemant.singh@unsw.edu.au
}

\SetAuthorBlock{Tapabrata Ray}{%
School of Engineering and Technology,\\
The University of New South Wales,\\
Canberra, Australia \\
email: t.ray@unsw.edu.au
}


\title{A Confidence-Driven Evolutionary Algorithm for Noisy Optimization with Joint Chance Constraints}




\begin{abstract}
Many real-world optimization problems involve noisy objective evaluations and probabilistic constraints, particularly in the form of joint chance constraints, which are computationally expensive to evaluate. In this work, we propose CR-EA-C, a confidence-driven evolutionary algorithm for solving noisy black-box optimization problems under joint chance constraints. CR-EA-C introduces three key components: (1) analytical feasibility estimation for joint chance constraints, (2) a pairwise statistical ranking mechanism for robust comparison under noise, and (3) a modified infeasibility-driven survival strategy to accelerate convergence. These components enable statistically reliable decision-making while improving the efficiency of function evaluations. The proposed method is evaluated against four recent metaheuristic algorithms under various uncertainty distributions. Furthermore, its practical effectiveness is also assessed on two additional real-world optimization problems and compared with conventional static sampling methods. Experimental results show that CR-EA-C consistently satisfies the prescribed joint chance constraints while achieving competitive objective values overall. This demonstrates that CR-EA-C is an effective general-purpose approach for noisy optimization.

\end{abstract}


\date{\today}

\maketitle

\section{Introduction and Related Work}

Real-world engineering design problems often require optimizing one or more performance objectives, subject to multiple constraints. A wide array of optimization algorithms exists, ranging from mathematical programming to metaheuristics. Metaheuristics are particularly practical when objective or constraint functions are highly nonlinear or lack an explicit analytical form, commonly referred to as \emph{black-box} problems. Although design optimization is a widely studied topic, majority of studies and approaches assume a deterministic setting. However, real-world systems are frequently subject to uncertainty arising from changing environmental conditions, manufacturing tolerances, stochastic simulations, and similar sources. Broadly speaking, these uncertainties manifest in three forms. The first is \textit{variable uncertainty}, often addressed within the domain of robust optimization~\cite{variable_uncertainty}, which accounts for imprecision or variability in decision inputs. The second is \textit{noisy optimization}, which concerns stochasticity in the objective function itself~\cite{obj_uncertainty}. Here, repeated evaluations of the same variable yield different outcomes. The third is \textit{chance-constrained optimization} (or probabilistic constrained optimization), in which feasibility must be guaranteed with a specified probability, reflecting the inherent randomness in constraint evaluation~\cite{Cooper59, newer_ref}. 

Within chance-constrained optimization, a further distinction is made between \textit{individual} and \textit{joint} chance-constrained problems (JCCPs)~\cite{types_of_chance_constraint}. Individual chance-constrained problems require each stochastic constraint inequality to be satisfied with its own probability, whereas joint chance-constrained problems demand that all constraints be met simultaneously with a certain probability~\cite{joint_chance_definition}. The latter presents a significantly greater challenge, as multiple random variables interact within a single feasibility condition.

A growing body of work has also examined chance-constrained combinatorial optimization problems, where decision variables are discrete and feasibility is defined probabilistically. Representative examples include a single chance-constrained variants of the knapsack problem~\cite{Neumann_Knapsack}, scheduling problems~\cite{neumann_scheduling}, the traveling thief problem~\cite{neumann_thief}, and submodular maximization~\cite{neumann_submodular}. These studies typically focus on a chance constraint that bounds the probability of violating a resource or capacity limit.

In this study, we focus on a continuous noisy black-box optimization problems subject to joint chance constraints, which have been scarcely studied so far. As previously discussed, for such problems in a black-box setting, the exact evaluation of the noiseless objective and constraint functions is generally not viable. More formally, the problem can be defined as follows~\cite{ASIA}. Let \(D\subset\mathbb{R}^p\) be a bounded $p$ dimension variable domain and let \(\xi\in\mathbb{R}^q\) be a  $q$ dimensional random vector with unknown distribution. For a reliability level \(\alpha\in(0,1)\), we seek a solution
\[
x^* \;=\;\arg\min_{x\in D}\;\mathbb{E}\bigl[f(x,\xi)\bigr]
\]
subject to both chance and deterministic constraints:
\begin{equation}
\begin{aligned}
\Pr\{G_i(x,\xi)\le0\}&\ge1-\alpha,\\
g_j(x)\,&\le0.\\
\end{aligned}
\label{eq:jccp}
\end{equation}
where $f$ is the stochastic objective function,
$G_i$ is the \(i\)-th stochastic constraint function, and $g_j$ is \(j\)-th deterministic inequality constraint function.

A solution \(x\) is considered feasible if it satisfies all the constraints in equation \eqref{eq:jccp}.  To have a unified measure of infeasibility level, the constraint violation (CV) is defined as
\begin{align}
\Gamma(x)
= \max\!\bigl\{\,1-\alpha - p(x),\,0\bigr\} 
\quad+\;\sum_{j=1}^J\max\!\bigl\{g_j(x),\,0\bigr\},
\label{eq:cv}
\end{align}
where
$p(x)\;=\;\Pr\bigl\{\,G(x,\xi)\le0\bigr\}$.

Note that \(\Gamma(x)=0\) if and only if \(x\) meets all constraints, including  chance and deterministic types.

\subsection{Strategies to Solve JCCP}
Due to the inherent intractability of joint chance-constrained problems, which are convex only under log-concave distributions, most practical solution methods rely on approximation schemes~\cite{important_cotributions}. Two common strategies are sampling-based approximations and analytical approximations. 

Sampling-based methods estimate the chance constraint’s underlying uncertainty distribution by drawing multiple independent samples. Some of the earliest work in this area include \cite{Pagnoncelli,Luedtke}. The main drawback of sampling-based approaches is that they typically require very large sample sizes to guarantee high-confidence feasibility, leading to substantial computational burden and limiting practical scalability.

In comparison, analytical approximation methods address chance constraints through a different mechanism. They use sampled data to replace stochastic constraints with deterministic formulations derived from probabilistic inequalities, such as Chebyshev’s inequality, Bernstein’s inequality, and Hoeffding’s inequality \cite{important_cotributions}.

One of the key challenge for both methods is deciding how many samples need to be allocated for each candidate solution, without overspending computational effort to obtain reliable response estimates. A widely used concept in stochastic simulation optimization is Optimal Computing Budget Allocation (OCBA) \cite{OCBA_survey}. In noisy optimization settings, OCBA approach helps to allocate evaluation budget among solutions to maximize the probability of correctly identifying the best one \cite{OCBA_def}. By focusing trials on the most competitive designs and  reducing their approximation variance, OCBA minimizes wasted effort on clearly inferior alternatives. 

Given the focus of this paper on black-box problems with noisy objective and JCCP, we discuss some of the recent metaheuristic approaches that tackle it here. These approaches are also used later in the study for benchmarking. A notable recent approach is the Adaptive Sampling Immune Algorithm ~(ASIA)~\cite{ASIA}. Empirical studies show that ASIA outperforms or matches the performance of other peer algorithms such as HPSO~\cite{HPSO}, SSGA-I/II~\cite{SSGA}, and IOM~\cite{IOM} across a set of test cases. Each of these approaches have distincet strategies for handling stochasticity and feasibility under chance constraints.

HPSO~\cite{HPSO} employs a simulation-based surrogate modeling approach, where random simulations are first used to generate training data for a BP neural network. This surrogate is then embedded within a Particle Swarm Optimization (PSO) framework to approximate fitness values and evaluate feasibility. However, BP neural networks generally require a relatively large amount of training data to achieve good predictive accuracy. In computationally expensive optimisation problems, where the evaluation budget is limited, generating sufficient training samples may itself become impractical.

SSGA~\cite{SSGA} adopts a steady-state genetic algorithm. SSGA-I and SSGA-II is a variant of their optimality scoring rules during fitness evaluation. Optimality scoring quantifies the relative quality of an individual’s objective value with respect to the best solution in the current population.  The algorithms evaluate the fitness of a solution by combining its feasibility score, \(v_*(v)\), and optimality score, \(\theta(v)\), as
\begin{equation}
\Psi_{\lambda}(v_*(v),\theta(v))
=
v_*(v)^{1-\lambda_m}\theta(v)^{\lambda_m}.
\end{equation}
where \(\lambda_m\) controls the relative emphasis on optimality and feasibility. Initially, the algorithms start with \(\lambda_m=0.99\), prioritising optimality. Hence, in early search stages, infeasible solutions are deliberately tolerated. As the search progresses, \(\lambda_m\) decreases towards \(0.01\), progressively increasing the emphasis on feasibility. 

While this strategy encourages broad exploration during the early search, an alternative approach that prioritises identifying feasible regions first and shifts greater weight to objective quality only after solutions enter the feasible region may be more suitable for computationally expensive noisy optimisation with a limited evaluation budget. By locating feasible regions earlier, the algorithm can reduce evaluations spent on high-quality but infeasible solutions and allocate more of the remaining budget to objective optimisation within the feasible space.

IOM~\cite{IOM} can be viewed as a conceptual predecessor to ASIA. It is a simple immune algorithm developed to handle non-joint chance-constrained problems, whereas ASIA is specifically designed to handle joint chance-constrained problems. IOM employs a sampling scheme in which all empirically feasible individuals in the current population are assigned the same larger sample size, while empirically infeasible individuals are evaluated using a smaller sample size. 

ASIA integrates analytical approximations with an adaptive sampling scheme and a modified ranking procedure within an immune-inspired algorithm  \cite{ASIA}. Throughout the search, feasibility is assessed using $\hat{p}(x)$, the lower-bound confidence estimate of $p(x)$, rather than the true $p(x)$. Solutions are then ranked according to the principle of reliability-dominance~\cite{Reliability-dominance}, where a solution $x$ is considered to dominate solution $y$ if one of the following conditions holds:
\begin{itemize}
    \item $x$ and $y$ are feasible and $\mathbb{E}[f(x,\xi)] < \mathbb{E}[f(y,\xi)]$
    \item $x$ is feasible and $y$ is infeasible
    \item $x$ and $y$ are infeasible and $\Gamma(x)<\Gamma(y)$
\end{itemize}

Then, to distribute the sampling effort adaptively by assigning more samples to better solutions, ASIA introduces the following equations:
\begin{align}
M_n &= \mathrm{round}\bigl(m_0\,N\sqrt{1 + n}\bigr),
\label{eq:Mn}\\
d(x) &= \bigl|\{\,y\in X \mid x \prec y\}\bigr|,
\label{eq:dx}\\
n(x) &= \mathrm{round}\!\Biggl(\frac{M_n\,d(x)}{\sum_{y\in X}d(y)}\Biggr).
\label{eq:nx}
\end{align}
where
\begin{description}
  \item[$m_0$] is a fixed parameter,
  \item[$N$] is a population size,
  \item[$n$] is a current generation index,
  \item[$M_n$] is a total sample budget at generation \(n\),
  \item[$X$] is the set of candidate solutions,
  \item[$x\prec y$] is the reliability‐dominance relation between \(x\) and \(y\),
  \item[$d(x)$] is the number of solutions dominated by \(x\),
  \item[$n(x)$] is the number of samples assigned to \(x\) in this generation.
\end{description}

\subsection{Motivation and contributions}
While existing evolutionary frameworks for noisy optimization have achieved encouraging results, several opportunities remain unexplored to further enhance their efficiency. First, sampling strategy can be improved by dynamically adjusting the generation-level sampling budget using OCBA mechanism that allocates additional evaluations only where they are most beneficial for identifying the optimum, rather than automatically assigning more samples to the current best-performing solution. Second, inspired by the infeasibility-driven survival mechanism proposed in \cite{IDEA}, a new survival strategy can be introduced to better preserve near-boundary solutions and guide the search towards the optimal region(s).

Building upon these insights, the motivation of this study is to extend the capability of Confidence Ranking Evolutionary Algorithm (CR-EA) \cite{paper1}, a recent general-purpose algorithm developed for unconstrained noisy optimization, to address noisy optimization problems subject to joint chance constraints. Towards this goal, we present CR-EA-C, which introduces two major extensions to handle probabilistic constraints effectively. First, CR-EA-C employs analytical feasibility approximations based on the Clopper–Pearson method. Second, it integrates a modified infeasibility-driven survival strategy inspired by Infeasibility Driven Evolutionary Algorithm (IDEA)~\cite{IDEA}. 

The Clopper–Pearson interval~\cite{Clopper-Pearson} is an exact two-sided confidence bound derived using the equal-tail rule, providing conservative estimates of a binomial proportion. A binomial proportion refers to the probability of success in a sequence of independent Bernoulli trials, estimated as the ratio of the number of successful outcomes to the total number of trials. In the joint chance-constrained setting, each evaluated solution can be classified as either \textit{feasible} or \textit{infeasible}, allowing the true feasibility probability $p(x)$ to be modeled as a binomial proportion. Reliable estimation of $p(x)$ requires avoiding overestimation, even at the cost of mild conservatism. A recent work~\cite{Clopper-Pearson_comparisson} compares four methods for constructing confidence intervals of binomial proportion, namely the Wald, Wilson score, Clopper–Pearson, and Likelihood methods. Through extensive empirical evaluation, the study demonstrates that the Clopper–Pearson interval consistently produces wider confidence intervals than the other methods across various sample sizes. This conservatism makes it particularly suitable for maintaining strict bounds on $p(x)$. Hence, it is a suitable choice to be paired with the proposed CR-EA-C.

IDEA \cite{IDEA} is an evolutionary algorithm that retains a small proportion infeasible solutions to explore constraint boundaries from both feasible and infeasible regions. This has shown to improve convergence, especially for the cases with constricted or disconnected feasible regions. The motivation for adopting this concept in joint chance-constrained optimization is not only to gain the benefit of approaching the constraint boundary from both sides of the search space, but also to preserve potentially good solutions with uncertain feasibility from dying prematurely, allowing them to receive sufficient samples before being discarded.

We compare the performance of CR-EA-C with multiple algorithms using the benchmark data provided in \cite{ASIA}. Experimental results demonstrate that CR-EA-C consistently satisfies the prescribed joint chance constraints while achieving superior objective values compared to the other algorithms.

Section~II provides an overview of the baseline algorithm, CR-EA. Section~III presents the proposed algorithm in detail, followed by the experimental setup in Section~IV. Benchmark results are reported and discussed in Section~V. Section VI presents ablation and parametric study. Additional experiments on real-world test problems are conducted in Section VII,  while concluding remarks and directions for future research discussed in Section~VIII.

\section{Overview of CR-EA}
CR-EA \cite{paper1} is a general-purpose evolutionary algorithm (EA) for unconstrained single-objective noisy optimization, designed to achieve improved sample efficiency under both homoscedastic and heteroscedastic noise. It combines a statistically grounded ranking mechanism with adaptive resampling and a modified tournament selection scheme, enabling robust discrimination between competing solutions under limited evaluation budgets. At its core, CR-EA employs Welch’s $t$-test for pairwise comparisons under unequal variances, while OCBA-inspired principles guide selective allocation of additional samples to solutions that are most informative for identifying the optimum. Through this tightly integrated design, CR-EA demonstrated strong performance across a wide range of noisy black-box benchmark problems in \cite{paper1}, making it a natural baseline for sample-efficient noisy optimization and an ideal foundation for exploring principled noise-handling mechanisms in constrained and risk-aware settings.

Enabling CR-EA to handle joint chance-constrained problems under noisy objective evaluations requires substantive extensions, which is the focus of this study. Unlike the unconstrained setting, candidate solutions must be assessed with respect to both objective quality and probabilistic feasibility, requiring simultaneous and reliable estimation of both quantities. Since both mechanisms rely on the same limited function evaluation budget, allocating resources to improve objective quality inevitably reduces the resources available for improving confidence in probabilistic feasibility, and vice versa. The algorithm must therefore balance exploration of promising but potentially infeasible regions with exploitation of solutions that are confidently feasible yet possibly suboptimal. Moreover, integrating OCBA becomes substantially more complex, as it is no longer evident whether additional samples should be allocated to reduce uncertainty in feasibility estimation or to refine objective ranking among feasible candidates.

\section{Proposed Algorithm}

The core idea of the algorithm is to concentrate sampling effort on solutions with promising objective values 
whose feasibility remains uncertain. We divide the evolutionary process into two stages:
\begin{enumerate}
  \item \textbf{Stage 1: Searching for the Feasible Region.}  
    In this stage, we emphasize identifying feasible regions of the search space.
  \item \textbf{Stage 2: Exploring the Feasible Region.}  
    Once at least $\tau$ solutions in the population are flagged as feasible, we shift focus to refining their objective values. The threshold $\tau$ is chosen to align this transition with the sampling budget adaptation mechanism described later in Section~\ref{increase_sample_size}.
\end{enumerate}

\begin{figure*} [!htbp]
  \includegraphics[width=0.98\textwidth]{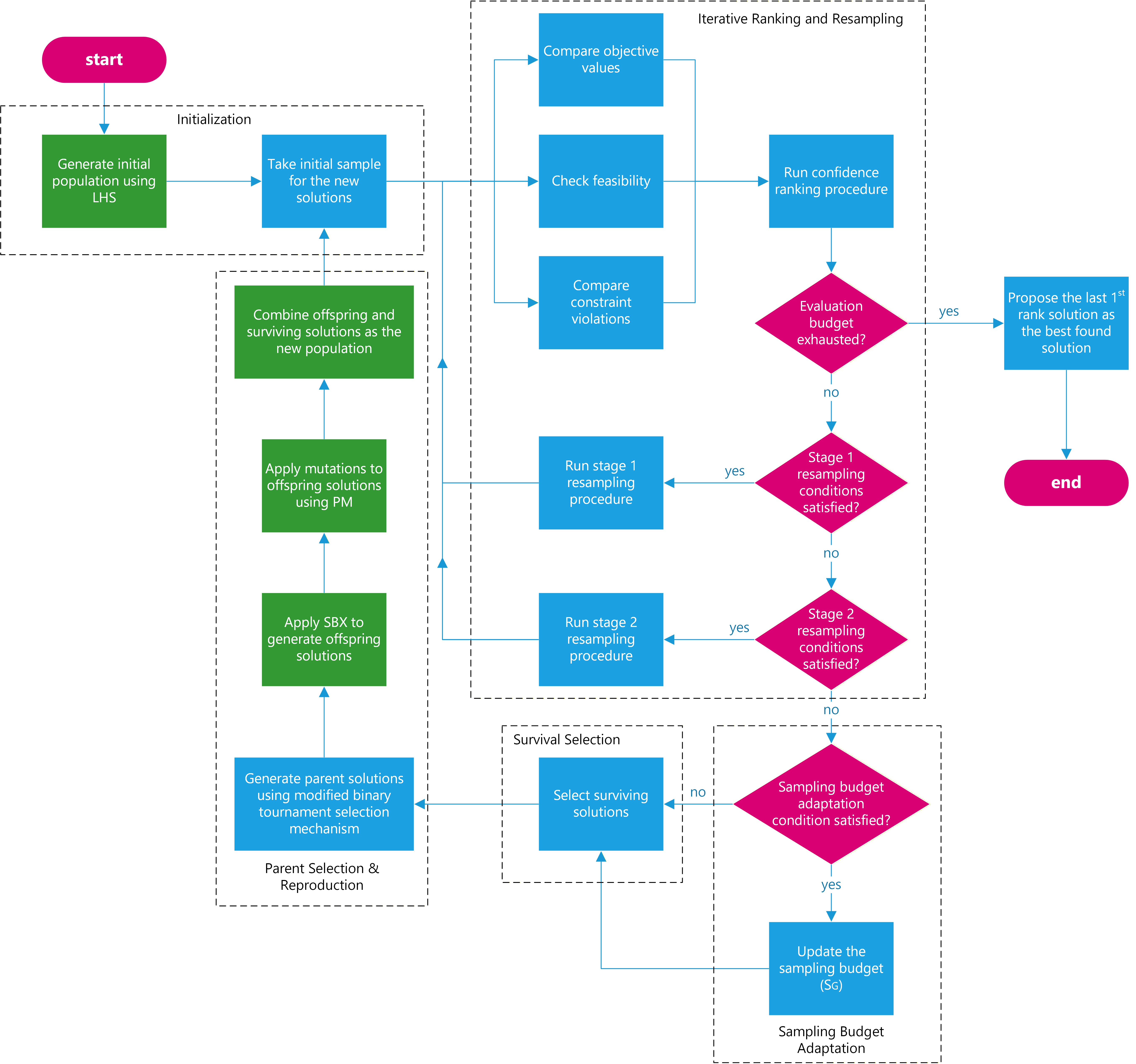}
\caption{Flowchart of CR-EA-C. The green boxes correspond to standard EA components, whereas the other elements correspond to the novel mechanisms introduced in this work.}
  \label{fig:flowchart_CR-EA-C}
\end{figure*}

To give a better holistic view of the algorithm, a flowchart is also provided in Figure~\ref{fig:flowchart_CR-EA-C}.

\subsection{Initialization}
To initialize, we set the population size to \(N = 10D\), where \(D\) is the decision variable dimension. An initial pool of \(N\) candidate solutions is generated via Latin hypercube sampling (LHS). We then compute the minimum sample size required (\(S_i\)) to determine feasibility by calculating how many samples are needed to raise the Clopper–Pearson lower confidence bound of \(p(x)\) above \(1-\alpha\) if the underlying $p(x) = 1$. Finally, the total sampling budget ($S_G$) that is allocated to each generation is calculated by equation \ref{eq:SG}.
\begin{equation}
S_G = 2N \times S_i
\label{eq:SG}
\end{equation}

\subsection{Iterative Ranking and Resampling}

Each new candidate solution is first evaluated  $s_0$ times. Thereafter, objective and constraints are used in ranking the solutions as follows.
\medskip
\subsubsection{\textbf{Comparing Objective Value}}
 We then compare every pair of solutions’ mean objective value using Welch’s $t$–test at 95\% confidence level. When solution $x$ is significantly better than its peer $y$, $x$ receives an $f_{\mathrm{win}}$ score~(label). If it is significantly worse, it receives an $f_{\mathrm{lost}}$ score. Otherwise, it is assigned an $f_{\mathrm{tie}}$ score.
\medskip
\subsubsection{\textbf{Checking Feasibility}}
We assess each solution’s feasibility by computing a two‐sided 99\% Clopper–Pearson confidence interval for its feasibility probability \(p(x)\). If the lower confidence bound ($LCB$) satisfies
\[
\mathrm{LCB}\bigl(p(x)\bigr) \;\ge\; 1 - \alpha,
\]
the solution is flagged as \emph{`feasible'}. If the upper confidence bound ($UCB$) satisfies
\[
\mathrm{UCB}\bigl(p(x)\bigr) \;<\; 1 - \alpha,
\]
it is flagged as \emph{`infeasible'}. Otherwise, it is flagged as \emph{'maybe feasible'}. Once a solution has been flagged as \textit{`feasible'}, it is treated as permanent, i.e., the tag is not overwritten during the search.

\medskip
\subsubsection{\textbf{Comparing Constraint Violations}}
For every sample of \(x\), we compute the sample constraint violation as:

\begin{align}
CV(x)
&=\;\sum_{j=1}^J\max\!\bigl\{G_i(x,\xi),\,0\bigr\}\nonumber+\;\sum_{j=1}^J\max\!\bigl\{g_j(x),\,0\bigr\} \nonumber.
\label{eq:cv_internal}
\end{align}
Note that this is different from the aggregate constraint violation in \eqref{eq:cv}.

We then compare every pair of solutions using Welch’s \(t\)–test at the 95\% confidence level on their mean \(\mathrm{CV}\). If solution \(x\) is significantly better (lower CV) than its partner \(y\), it receives a \(CV_{\mathrm{win}}\) score. If it is significantly worse, it receives a \(CV_{\mathrm{lost}}\) score. Otherwise, it is assigned a \(CV_{\mathrm{tie}}\) score.
\medskip
\subsubsection{\textbf{Ranking Procedure}}
Solutions are first ordered by their feasibility flag, with \emph{feasible} solutions ranked better than \emph{maybe} solutions, which in turn are ranked better \emph{infeasible} ones. Within each feasibility tier, we apply a lexicographic ranking according to the following rules:

\medskip

\noindent\textbf{Feasible:} sort by the total numbers of \[
\bigl(f_{\mathrm{win}},\,f_{\mathrm{tie}},\,-f_{\mathrm{lost}}\bigr)
\]
in descending lexicographic order~(i.e.,  higher value is better).

\medskip

\noindent\textbf{Maybe (Stage 1):} In Stage 1, first sort by the Clopper–Pearson lower confidence bound in descending order. Ties are broken by the total number of 
\[
\bigl(CV_{\mathrm{win}},\,CV_{\mathrm{tie}},\,-CV_{\mathrm{lost}}\bigr),
\]
and then by the total number of
\[
\bigl(f_{\mathrm{win}},\,f_{\mathrm{tie}},\,-f_{\mathrm{lost}}\bigr),
\]
all in descending lexicographic order.

\medskip

\noindent\textbf{Maybe (Stage 2):} In Stage 2, the solutions are first sorted by
\[
\bigl(f_{\mathrm{win}},\,f_{\mathrm{tie}},\,-f_{\mathrm{lost}}\bigr),
\]
then by
\[
\bigl(CV_{\mathrm{win}},\,CV_{\mathrm{tie}},\,-CV_{\mathrm{lost}}\bigr),
\]
and finally by the Clopper–Pearson lower confidence bound, each comparison is in descending order.

\medskip

\noindent\textbf{Infeasible:} The infeasible solutions are sorted by the total numbers of
\[
\bigl(CV_{\mathrm{win}},\,CV_{\mathrm{tie}},\,-CV_{\mathrm{lost}}\bigr),
\]
and then by the total numbers of
\[
\bigl(f_{\mathrm{win}},\,f_{\mathrm{tie}},\,-f_{\mathrm{lost}}\bigr),
\]
all in descending lexicographic order.

\medskip

\subsubsection{\textbf{Resampling Procedure}}
After the initial ranking, resampling is performed based on the current stage:

\begin{itemize}
    \item \textbf{Stage 1:} Collect all \textit{`maybe feasible'} solutions that have not yet reached their maximum sample size \(S_x\).  Each such solution receives an additional \(S_{+}\) samples, after which the ranking is updated. This process repeats until no qualifying solutions remain.

    \item \textbf{Stage 2:} Resampling process takes place in two phases. In the first phase, we collect all \textit{`maybe feasible'} solutions that have not yet reached their maximum sample size \(S_x\), and whose objective value is better than the current 1st rank solution. Each such solution receives an additional \(S_{+}\) samples, after which the ranking is updated. This process repeats until no qualifying solutions remain. In the second phase, we similarly identify all \textit{`feasible'} or \textit{`maybe feasible'} that have not yet reached their maximum sample size \(S_x\), and whose objective value is `tie' compared to the current 1st rank solution. Each such solution receives an additional \(S_{+}\) samples, after which the ranking is updated. This process repeats until no qualifying solutions remain. 
    
    The motivation behind this two phase resampling mechanism is to progressively allocate additional function evaluations to `contender' solutions that have potential to become the best solution but whose estimated objective value or feasibility remains uncertain. The formal definition of a contender solution is provided in Section~\ref{increase_sample_size}.

\end{itemize}

The Optimal Computing Budget Allocation (OCBA) framework~\cite{OCBA_survey} suggests that allocating sampling effort in proportion to the standard deviations of competing alternatives increases the probability of correct selection in stochastic simulations. Building on this principle, the proposed approach allocates more samples to solutions with higher potential, defined as those exhibiting superior objective values and greater likelihood of feasibility, while assigning fewer samples to solutions with lower potential.

Each solution's maximum sample size \(S_x\) is defined as
\begin{equation}
  S_x = \max\!\Biggl[
    \frac{S_G}{2}\,\biggl(\frac{\sigma_x}{\sum_{y\in X}\sigma_y}
    \;+\;
    \frac{\omega_x}{\sum_{y\in X}\omega_y}\biggr),
    \;S_i
  \Biggr]
  \label{eq:Sx}
\end{equation}

where \(\sigma_x\) is the sample standard deviation of solution \(x\)’s objective value, \(\sum_{y\in X}\sigma_y\) is the total objective value's standard deviation for all solution in the current population, \(\omega_x\) is the half-width of Clopper–Pearson confidence interval of solution \(x\), \(\sum_{y\in X}\omega_y\) is the sum of those half-widths for all solution in the current population, and $S_i$ is the same as in equation \ref{eq:SG}.

The incremental sample allocation \(S_{+}\) is then calculated as
\begin{equation}
S_{+} \;=\; \max\!\Biggl(\bigl\lfloor \tfrac{S_x}{S_i}\bigr\rfloor, \;1\Biggr)
\label{eq:Sp_updated}
\end{equation}

\subsection{Survival Selection} \label{survival_selection}
At the end of each generation, the solution ranked first after the iterative ranking and resampling procedure is designated as the proposed best solution of that generation. 

To form the surviving population for the next generation, we first promote the top $0.2N$ \textit{`maybe feasible'} solutions to the highest rank. As this mechanism is inspired by IDEA \cite{IDEA}, we adopt the same parameter setting of $0.2N$. This setting also aligns naturally with the parameter $\tau=0.2N$, which controls the trigger of sampling budget adaptation described later in Section~\ref{increase_sample_size}.

From the reordered list, top $N$ solutions are then selected to survive into the next generation. This strategy maintains population diversity and mitigates premature convergence by allowing \textit{`maybe feasible'} solutions to continue contributing to the search. In doing so, it preserves exploratory potential around constraint boundaries, where the global optimum is often located~\cite{IDEA}. In addition, it prevents potentially good solutions~(those with promising objective values but undetermined feasibility) from being discarded prematurely, giving them an opportunity to be resampled again in later generations.

\subsection{Sampling Budget Adaptation} \label{increase_sample_size}
As discussed in \cite{paper1}, variations in noise across the search space can lead the population to incorrectly identify the top-ranked solution. To mitigate this effect, CR-EA employs a dynamic adjustment mechanism for the global sampling budget, allowing it to adaptively respond to the noise level inferred from the ranking process. When a significant portion of the population exhibits ties with the first-ranked solution by the end of a generation, it suggests that noise is strongly influencing ranking accuracy. 
In such cases, the global sampling budget is increased to enhance confidence in the selection outcome. 

\begin{figure}[!htbp]
    \centering
    \includegraphics[width=1\linewidth]{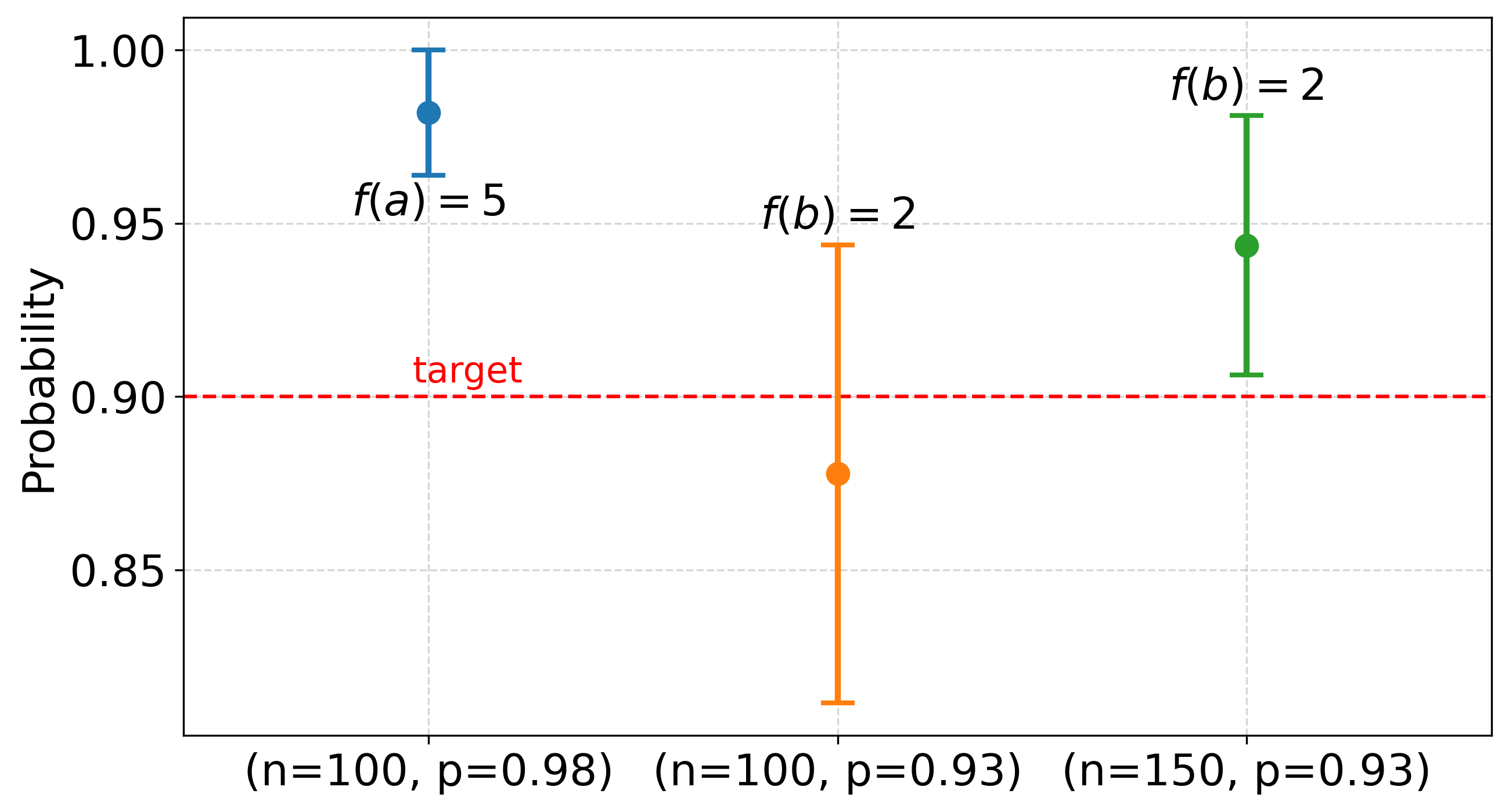}
    \caption{Affect of sampling budget adaptation.  $n$ is sample size and $p$ is probability of feasibility.}
    \label{fig:sampling_increase}
\end{figure}

To illustrate how insufficient sampling can distort decision-making in a joint chance-constrained setting, consider two known feasible solutions, \textit{a} and \textit{b} (Figure~\ref{fig:sampling_increase}). Solution \textit{a} has an objective value of 5 and a feasibility probability of 0.98, whereas solution \textit{b} has an objective value of 2 and a feasibility probability of 0.93. Since the goal is to minimize the objective value subject to a feasibility requirement of at least 0.90, solution \textit{b} is clearly superior to solution \textit{a}.

However, when the algorithm evaluates both locations using only 100 samples, the uncertainty in feasibility causes solution \textit{b} to be classified as \textit{maybe-feasible} (orange line). As a result, the algorithm incorrectly treats solution \textit{a} as the better candidate (blue line), despite its worse objective value.

When the number of samples is increased (green line), the feasibility estimate for solution \textit{b} becomes more accurate, allowing the algorithm to correctly identify it as feasible. Consequently, solution \textit{b} is now recognized as the superior solution. This example highlights the importance of adaptive resampling in joint chance-constrained optimization.

In this study, we modify the original approach by redefining the concept of a `tie' in CR-EA \cite{paper1} as a \textit{contender}. At the end of each generation, we count the number of contender solutions, defined as solutions that satisfy one of the following conditions:
\begin{enumerate}
    \item \textit{`feasible'} with objective‐value \textit{ties} to the first ranked solution,
    \item  \textit{`maybe‐feasible'} with objective‐value ties to the first ranked solution, or 
    \item \textit{`maybe‐feasible'} with better objective values than the first rank solution. 
\end{enumerate}
\noindent If the count of contender solutions exceeds $\tau$, we set $S_G=\beta \times S_G$ for the next generation.

\subsection{Parent Selection and Reproduction}
To reproduce, $N$ offspring are generated from $N$ parents using simulated binary crossover (SBX) and polynomial mutation (PM). Parents are chosen using a modified tournament selection that follows these rules: 
\begin{enumerate}
    \item  \textit{`Feasible'} solution is superior over \textit{`maybe feasible'} solution, which are in turn superior over \textit{`infeasible'} solution. 
    \item If both parents are \textit{`feasible'}, their mean objective values are compared using Welch’s \(t\)-test, lower is considered better.
    \item If both parents are \textit{`maybe feasible'}, the one with higher Clopper–Pearson lower confidence bound is considered better.
    \item If both parents are \textit{`infeasible'}, their mean constraint violation (CV) are compared using Welch’s \(t\)-test, lower is better. 
\end{enumerate}

\section{Experimental Design}
\subsection{Benchmark Problems}
We evaluate CR-EA-C on the three benchmark problems used in the recent study on the topic~\cite{ASIA}. Among these, one is a mathematical test problem, while the other two are derived from practical applications: an open storage network problem and an oil production planning problem. Throughout this study, the notation $\mathcal{N}(i,j)$ denotes a normal distribution with mean $i$ and variance $j$. The detailed problem formulations are provided in Appendix section~\ref{appendix:Benchmark Problems}.

\subsection{Parameter Settings}
\label{sec: parameter settings}
We perform 31 independent runs for each problem to enable selection of the median result (compared to the 30 runs reported in~\cite{ASIA}). Each run is allotted an identical budget of 150{,}000 function evaluations. The parameter settings of CR-EA-C are summarised in Table~\ref{tab:default-params}, where $D$ is problem dimension. 

Note that in~\cite{ASIA}, ASIA was evaluated with tuned parameter settings. Hence, to ensure a fair comparison, CR-EA-C is also evaluated using tuned parameters. In particular, the SBX crossover index is increased from 15 (in the original CR-EA \cite{paper1}) to 30. The corresponding motivation and parametric study for this modification are presented in Section~\ref{abla study}.

    \begin{table}[!ht]
        \centering
        \caption{Parameter settings for CR-EA-C}
        \label{tab:default-params}
        \begin{tabular}{lll}
            \toprule
            Parameter & Symbol                    & Value  \\
            \midrule
            Population size& $N$& $10D$    \\
            SBX crossover index& $\eta_c$& 30\\
            Probability of crossover& $P_c$& 1.0\\
            Polynomial mutation index& $\eta_m$& 20     \\
            Probability of mutation& $P_m$& 0.1    \\
            Minimum sample size& $s_0$& 3    \\
            Tie threshold to update budget & $\tau$ & 0.2$N$    \\
            Global sampling budget multiplier & $\beta$ & 2    \\
            \bottomrule
        \end{tabular}
    \end{table}

\section{Results and Discussion}
\subsection{Overview}
Table~\ref{tab:results_table} presents the results for all three benchmark problems. Here, AV denotes the average constraint violation, calculated via Equation~\ref{eq:cv}, and CI represents the $90\%$ confidence interval for the mean objective value across runs. The reported values are obtained by reevaluating each proposed final best solution of each run with 300{,}000 samples. To ensure a fair comparison, the best, worst, mean, standard deviation, Lower CI, and Upper CI values of CR-EA-C are computed using only the successful runs, with failed runs excluded from the statistics. However, it is unclear whether the results reported for the remaining algorithms in Table~\ref{tab:results_table} were computed using the same procedure in ~\cite{ASIA}.

The results indicate that CR-EA-C outperforms the other methods in Examples~1, 2 and~3. Although, in Example~1, four runs marginally failed to satisfy the joint chance constraint, yielding a nonzero AV. The proposed best solutions across these runs are just slightly below the 0.9 target probability threshold, yielding true $p(x)$ values of 0.896, 0.897, 0.898, and 0.899. Notably, three of the compared algorithms~(HPSO, SSGA-I, SSGA-II) have much higher AV, while two of them~(IOM,ASIA) have slightly lower AV but also slightly worse objective values than CR-EA-C. 

\begin{table*}[!htbp]
  \centering
  \caption{Performance summary for all algorithms on three problems \cite{ASIA}.}
  \label{tab:results_table}
  \begin{tabular}{@{} ll c c c c c c c @{}}
    \toprule
    \makecell{\textbf{Problem}} 
      & \makecell{\textbf{Algorithm}} 
      & \textbf{Best} & \textbf{Worst} & \textbf{Mean} & \textbf{STDEV}
      & \textbf{Lower CI} & \textbf{Upper CI} & \textbf{AV} \\
    \midrule
    \multirow{6}{*}{\makecell[c]{\textbf{Example 1}\\\footnotesize\emph{Open Storage Network (minimize)}}}
      & HPSO        &  64.085   & 109.347   &  73.883   & 10.476   &  71.431   &  76.335   & 0.0470   \\
      & SSGA-I      &  61.592   &  73.436   &  64.693   &  2.734   &  64.053   &  65.333   & 0.0530   \\
      & SSGA-II     &  61.898   &  67.721   &  64.420   &  1.659   &  64.032   &  64.808   & 0.0530  \\
      & IOM         & 129.973   & 180.823   & 147.552   & 12.268   & 144.680   & 150.423   & 0.0000  \\
      & ASIA        & 108.948   & 123.329   & 118.511   &  3.307   & 117.737   & 119.286   & 0.0000  \\
      & CR-EA-C & 105.692 & 131.212 & 116.7941 & 6.472 & 114.673 & 118.916 & 0.0004 \\
    \midrule
    \multirow{6}{*}{\makecell[c]{\textbf{Example 2}\\\footnotesize\emph{Oil Production Planning (minimize)}}}
      & HPSO        & 149.036   & 192.043   & 166.933   & 12.763   & 164.971   & 168.895   & 0.0014  \\
      & SSGA-I      & 142.069   & 155.298   & 144.281   &  3.128   & 143.800   & 144.762   & 0.0927  \\
      & SSGA-II     & 141.832   & 143.488   & 142.596   &  0.441   & 142.529   & 142.664   & 0.0949  \\
      & IOM         & 147.482   & 161.437   & 152.980   &  3.802   & 152.396   & 153.565   & 0.0000  \\
      & ASIA        & 145.986   & 148.632   & 147.616   &  0.566   & 147.529   & 147.703   & 0.0005  \\
      & CR-EA-C
                   & 136.735
                   & 141.688
                   & 138.457
                   & 1.3909
                   & 138.026
                   & 138.888
                   & 0.0000 \\
    \midrule
    \multirow{6}{*}{\makecell[c]{\textbf{Example 3}\\
    \footnotesize\emph{Multimodal Function Opt. (maximize)}}}
      & HPSO        &   1.992   &  -0.787   &   0.124   &  0.473   &   0.079   &   0.169   & 0.0112  \\
      & SSGA-I      &   9.581   &   8.320   &   9.101   &  0.331   &   9.069   &   9.133   & 0.0064  \\
      & SSGA-II     &   9.575   &   8.396   &   9.062   &  0.276   &   9.035   &   9.089   & 0.0045  \\
      & IOM         &   9.245   &   6.957   &   8.456   &  0.627   &   8.396   &   8.516   & 0.0000  \\
      & ASIA        &   8.919   &   7.498   &   8.598   &  0.338   &   8.566   &   8.631   & 0.0000  \\
      & CR-EA-C
                   & 12.844
                   & 9.319
                   & 11.750			
                   & 0.734
                   & 11.521
                   & 11.976
                   & 0.0000 \\
    \bottomrule
  \end{tabular}
\end{table*}

\subsection{CR-EA-C Behaviour}

\begin{figure*}[!h]
    \centering
    \includegraphics[width=1\linewidth]{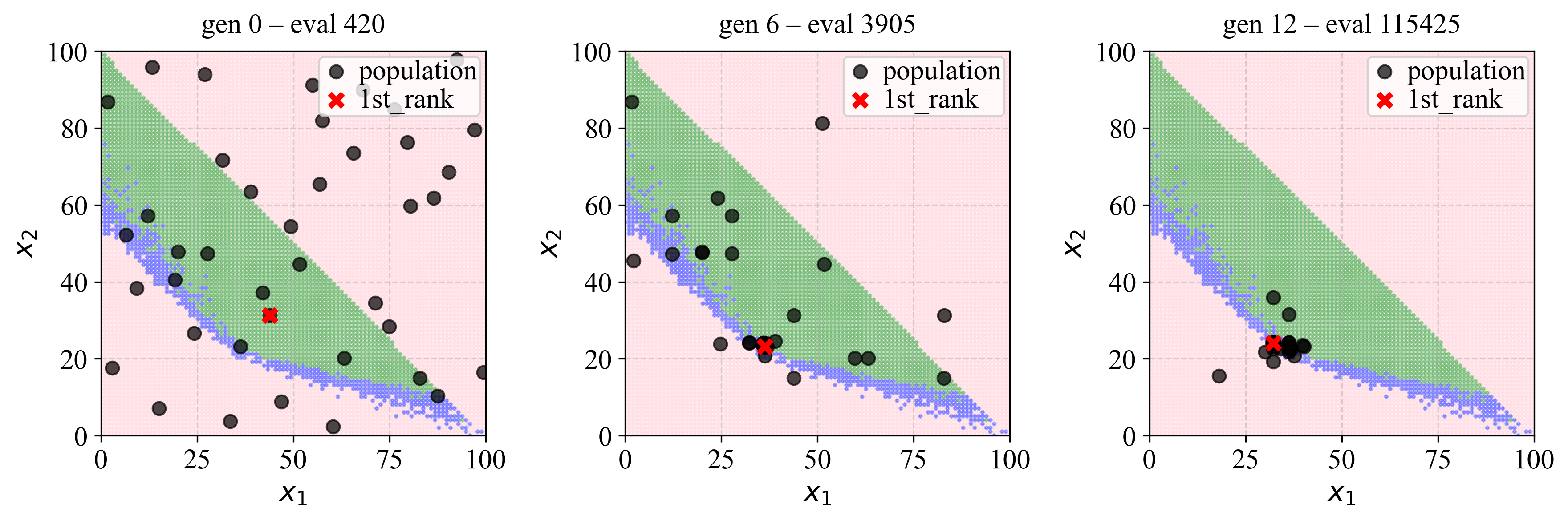}
    \caption{CR-EA-C median run's surviving population of Example 2.}
    \label{fig:moving_gen}
\end{figure*}

\begin{figure}[!h]
    \centering
    \includegraphics[width=1\linewidth]{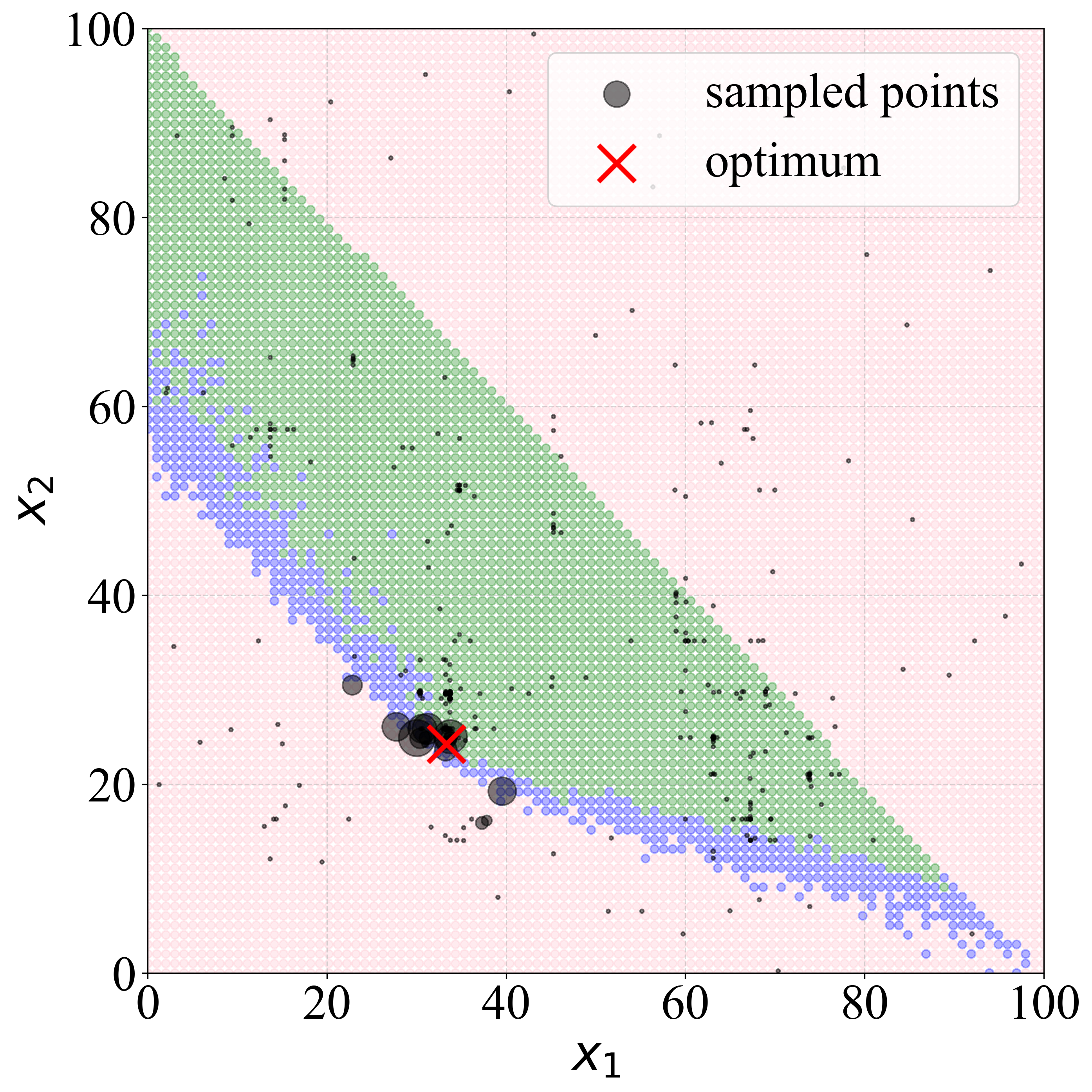}
    \caption{CR-EA-C median run's sampling visualization of Example 2.}
    \label{fig:sampling_size}
\end{figure}

To illustrate the search behavior of CR-EA-C, Figures~\ref{fig:moving_gen} and~\ref{fig:sampling_size} 
depict the search space of the two variable problem of Example~2. The data used in these figures are taken from the median run. Both figures employ green shading to indicate the feasible region, blue shading for the \textit{'maybe feasible'} region, and red shading for the infeasible region. These illustrations are based on 50 samples taken at each point. Figure~\ref{fig:moving_gen} shows the surviving population of CR-EA-C in the median run for Example~2. 
Red crosses mark the best solution selected at each generation. It can be observed that some infeasible solutions survive near the optimum, which helps guide the rest of the population more rapidly toward the optimum. Figure~\ref{fig:sampling_size} displays all unique solutions identified by CR-EA-C, with circle size proportional to the number of samples taken at each point. This plot demonstrates how CR-EA-C strategically allocates larger sampling budgets near the optimal region.

\subsection{True Rank Plot}

\begin{figure*}[!ht]
  \centering
  \subfloat[Example 1]
    {\includegraphics[width=0.32\textwidth]{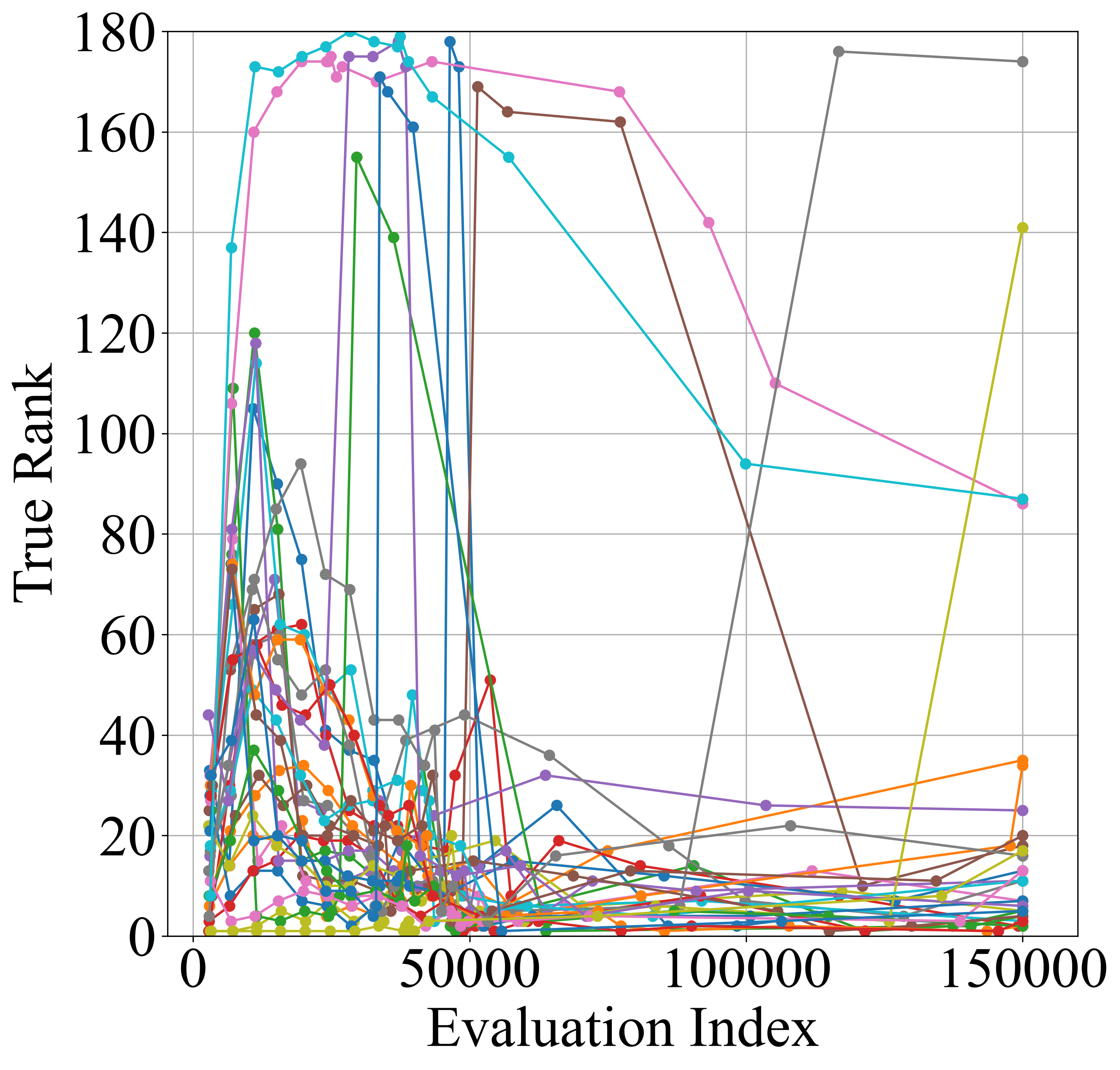}%
     \label{fig:true_rank_all_1}}
  \hfil
  \subfloat[Example 2]
    {\includegraphics[width=0.32\textwidth]{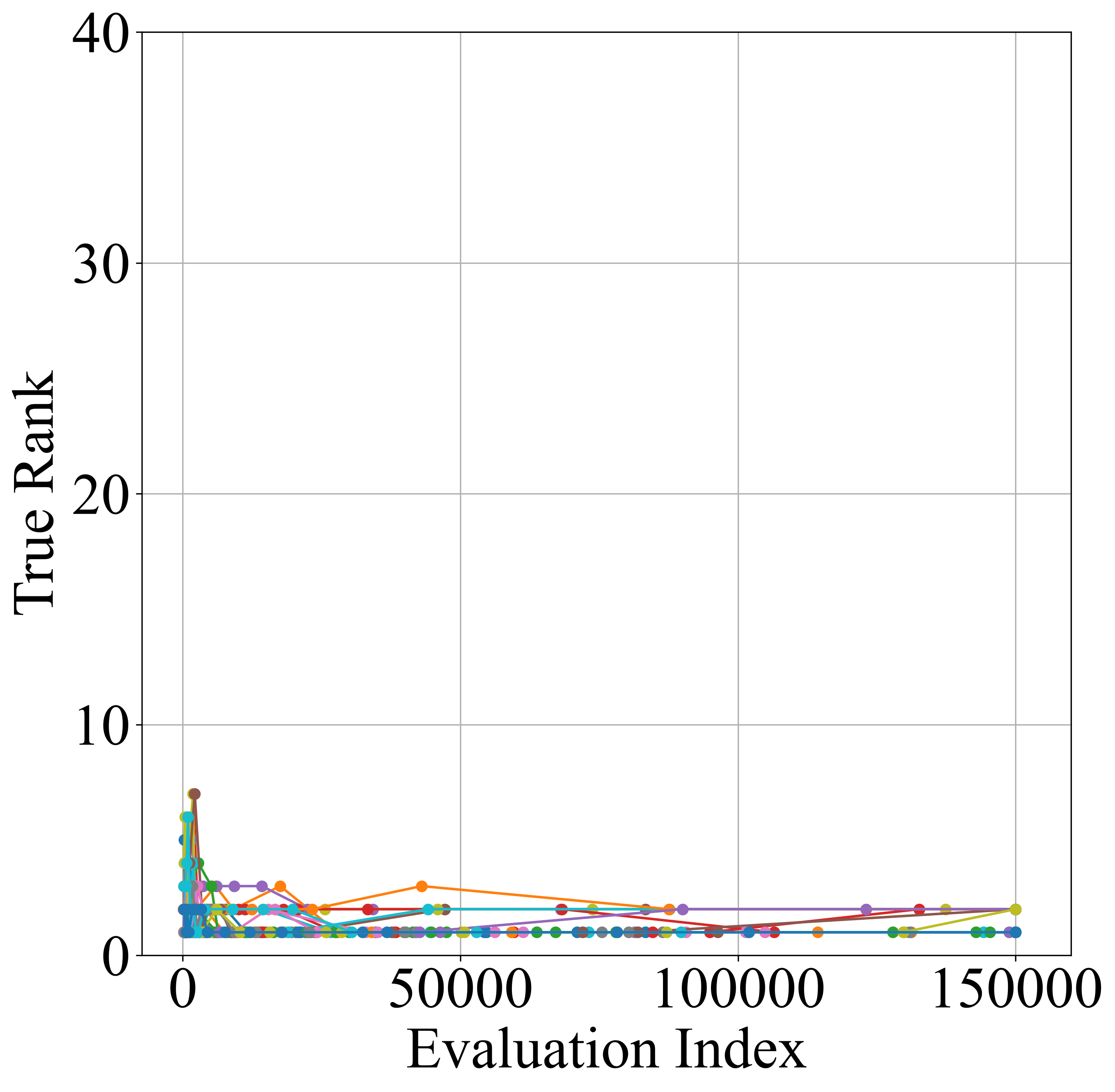}%
     \label{fig:true_rank_all_2}}
  \hfil
  \subfloat[Example 3]
    {\includegraphics[width=0.32\textwidth]{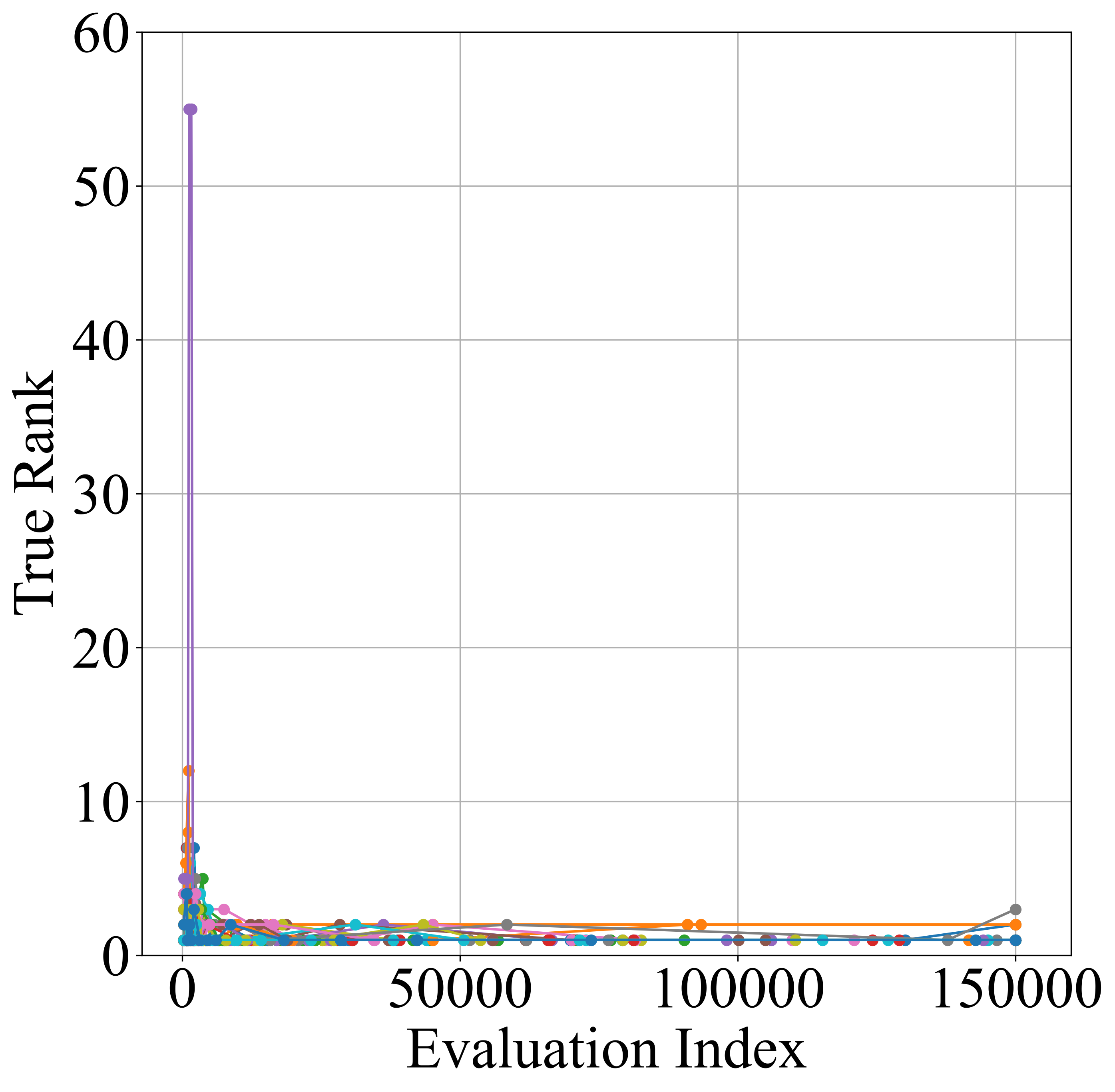}%
     \label{fig:true_rank_all_3}}
  \caption{True rank of CR-EA-C's first‐rank solution in each generation for Examples 1–3.}
  \label{fig:true_rank_all}
\end{figure*}

\begin{figure*}[!htbp]
  \centering
  \subfloat[Example 1]{%
    \includegraphics[width=0.32\textwidth]{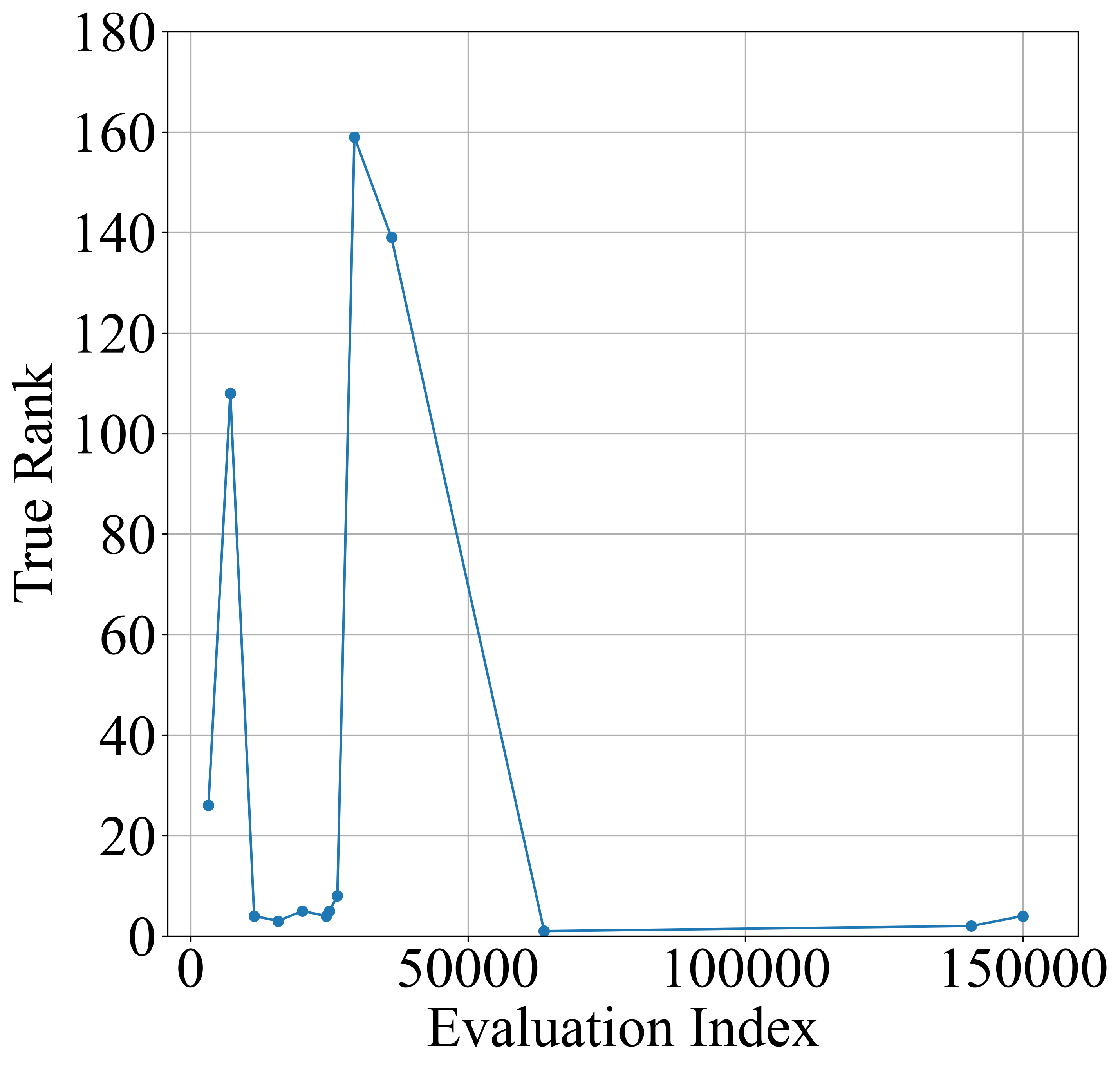}%
    \label{fig:true_rank_median_1}
  }%
  \hfil
  \subfloat[Example 2]{%
    \includegraphics[width=0.32\textwidth]{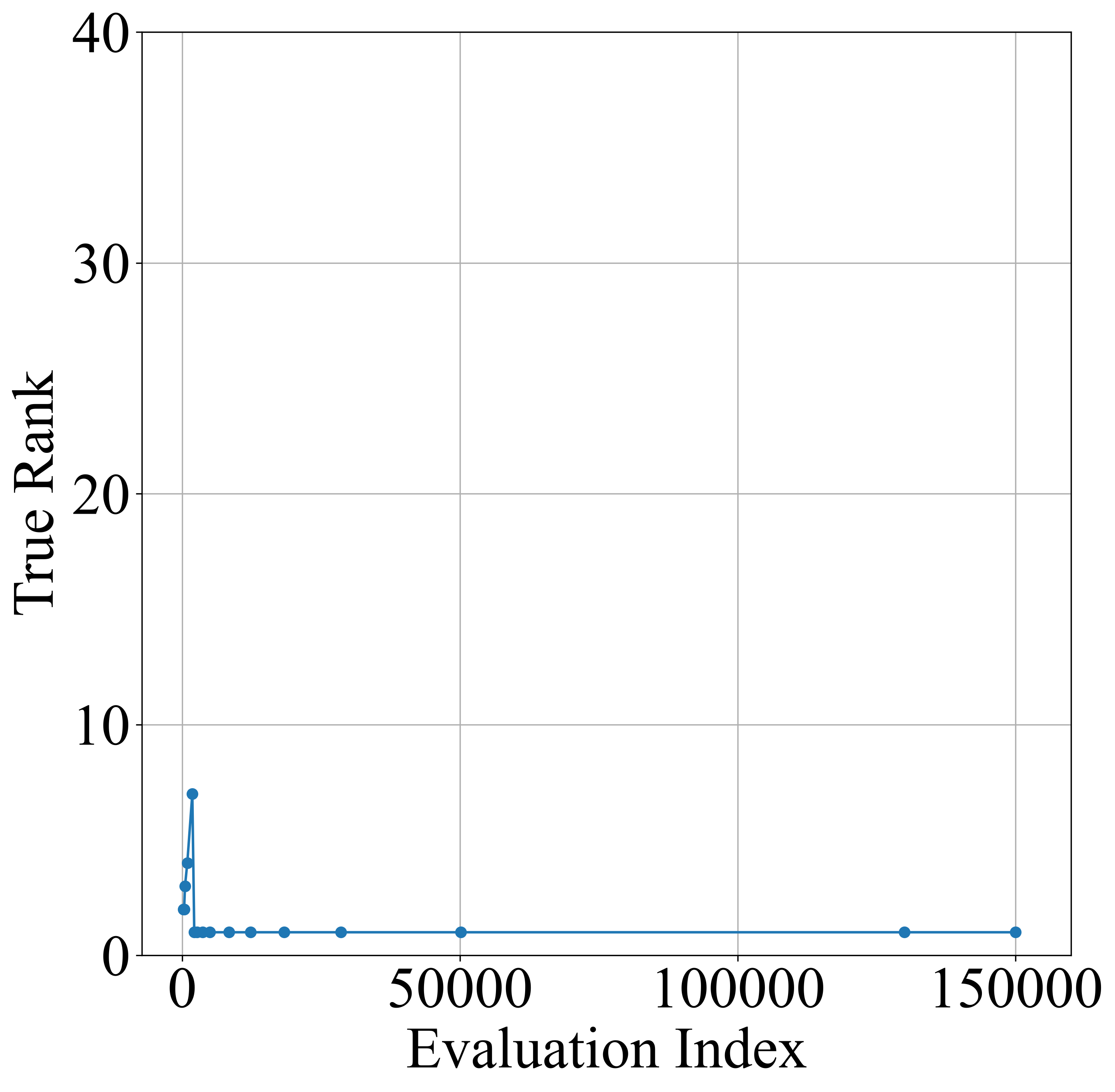}%
    \label{fig:true_rank_median_2}
  }%
  \hfil
  \subfloat[Example 3]{%
    \includegraphics[width=0.32\textwidth]{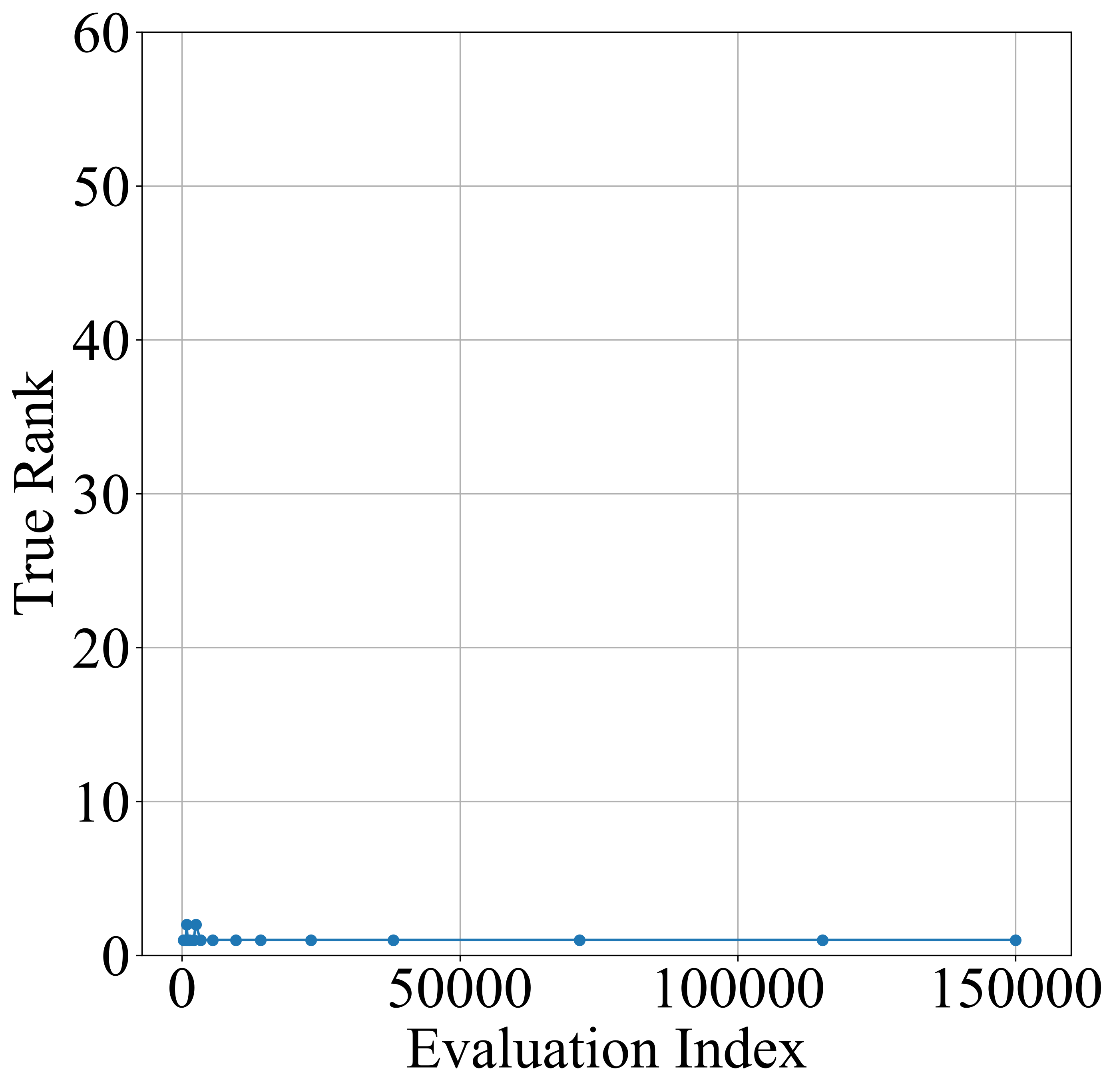}%
    \label{fig:true_rank_median_3}
  }%
  \caption{CR-EA-C's median run's true rank of the first‐rank solution in each generation for Examples 1–3.}
  \label{fig:true_rank_median}
\end{figure*}

Figures~\ref{fig:true_rank_all} and~\ref{fig:true_rank_median} show the true-rank of CR-EA-C's first-rank solution for all 31 runs and for the median run, respectively. Here, the true rank of a solution is defined as its rank when all candidate solutions in the population are evaluated using their true (noiseless) objective values and constraint values. Thus, the true rank of CR-EA-C's first-ranked solution indicates its actual position in the noiseless population, with a true rank of 1 representing the globally best solution within that generation. These plots therefore illustrate how accurately the algorithm identifies the best solution.

CR-EA-C generally selects strong solutions, though Example~1 exhibits greater variability due to its eight constraints and limited feasible region, which makes the algorithm more prone to misclassify infeasible points as feasible. Nevertheless, CR-EA-C is able to correct such errors over subsequent generations by encountering superior offspring. The final results reveal that four runs produced infeasible solutions, explaining the four points with high last values at  end of Figure~\ref{fig:true_rank_median_1}.

\section{Ablation and Parametric Study}
\label{abla study}

\subsection{Ablation Study}
As introduced in Sections~\ref{survival_selection} and~\ref{increase_sample_size}, CR-EA-C incorporates a \textit{`maybe feasible'} driven survival selection mechanism. To investigate the effect of this component, an ablation study was conducted using a modified version of CR-EA-C, referred to as CR-EA-C-FD, on the same problem over 31 independent runs. This variant replaces the original mechanism with a standard feasibility-driven (elitist) survival selection.

\begin{figure*}[!htbp]
  \centering
  \subfloat[Example 1]
    {\includegraphics[width=0.32\textwidth]{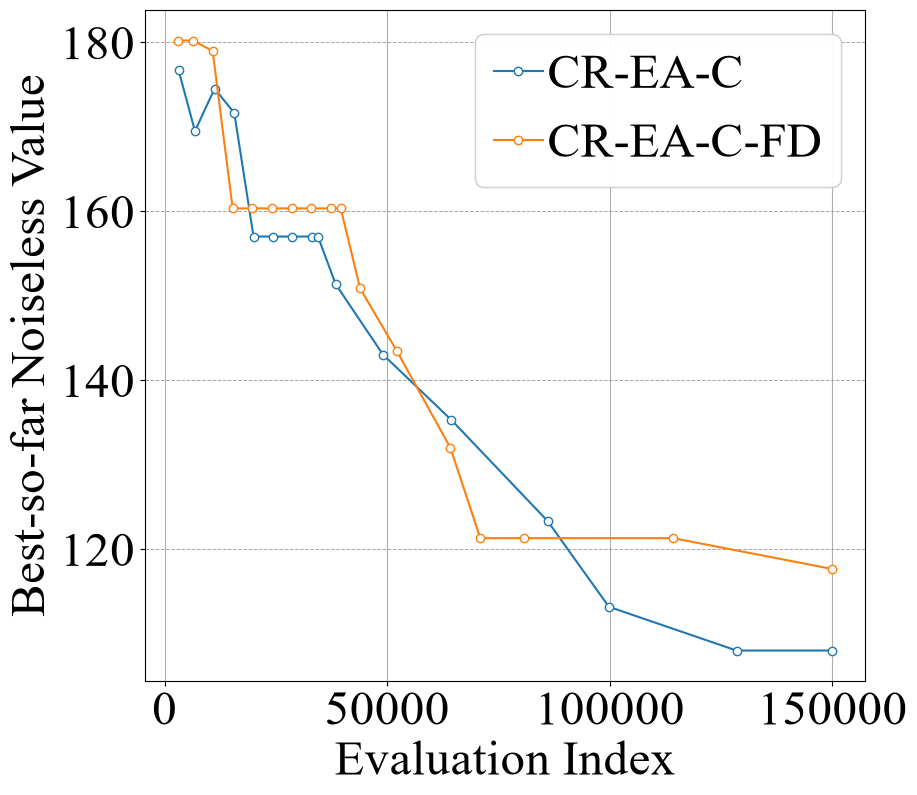}%
     \label{fig:convergence_median_1}}
  \hfil
  \subfloat[Example 2]
    {\includegraphics[width=0.32\textwidth]{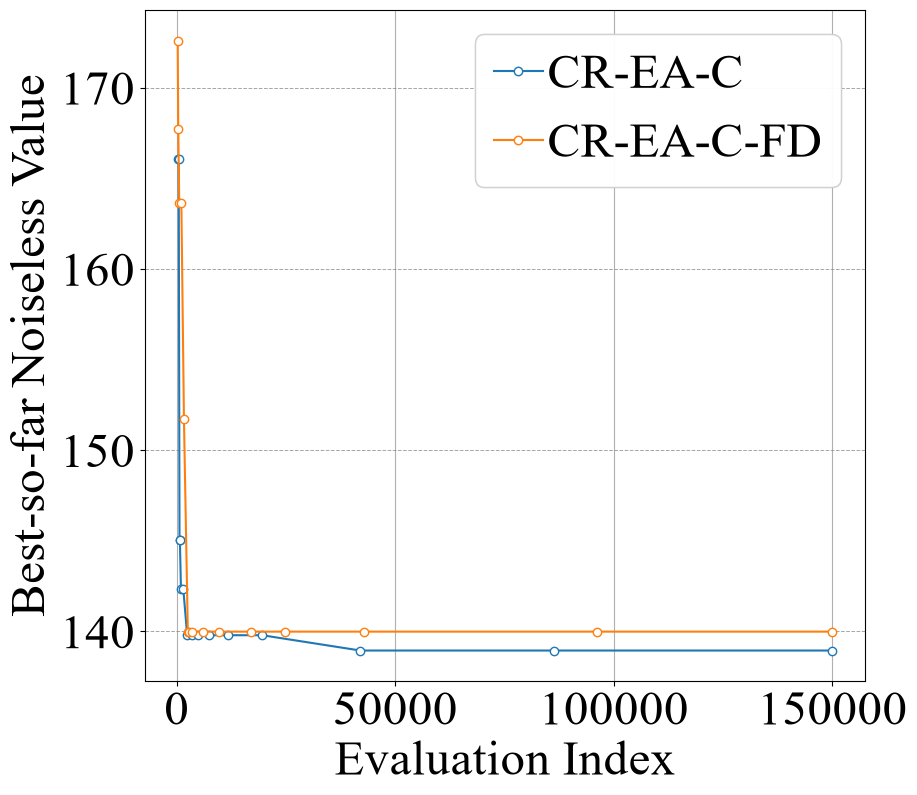}%
     \label{fig:convergence_median_2}}
  \hfil
  \subfloat[Example 3]
    {\includegraphics[width=0.32\textwidth]{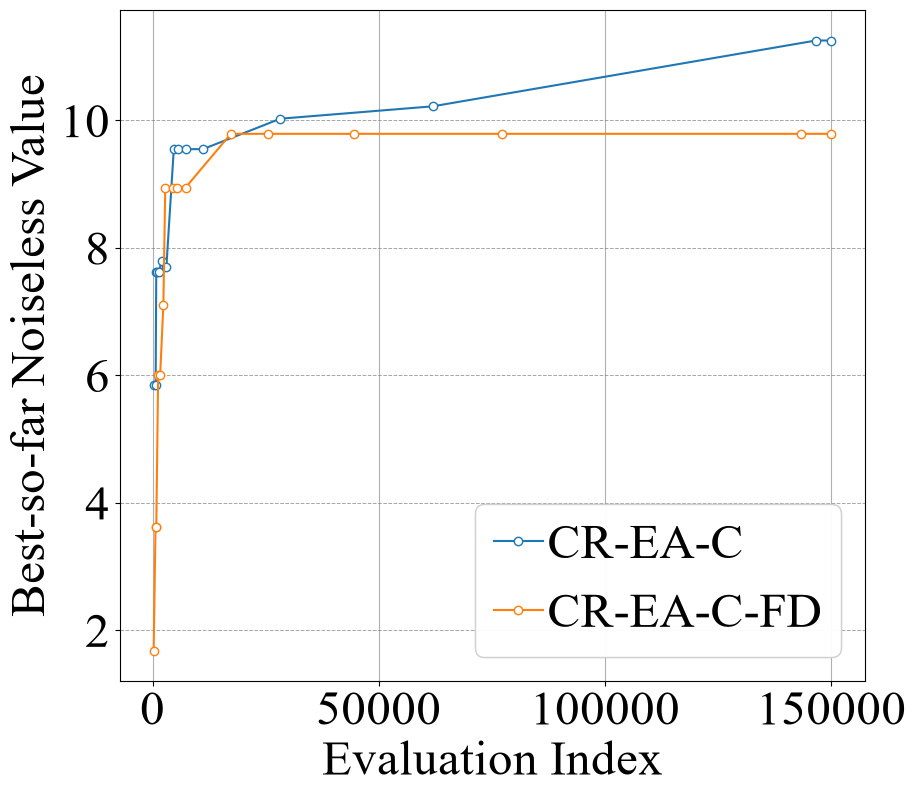}%
     \label{fig:convergence_median_3}}
  \caption{CR-EA-C's median run's first‐rank solution's noiseless value in each generation for Examples 1–3.}
  \label{fig:convergence_median}
\end{figure*}

\begin{figure*}[!htbp]
  \centering
  \subfloat[Example 1]
    {\includegraphics[width=0.32\textwidth]{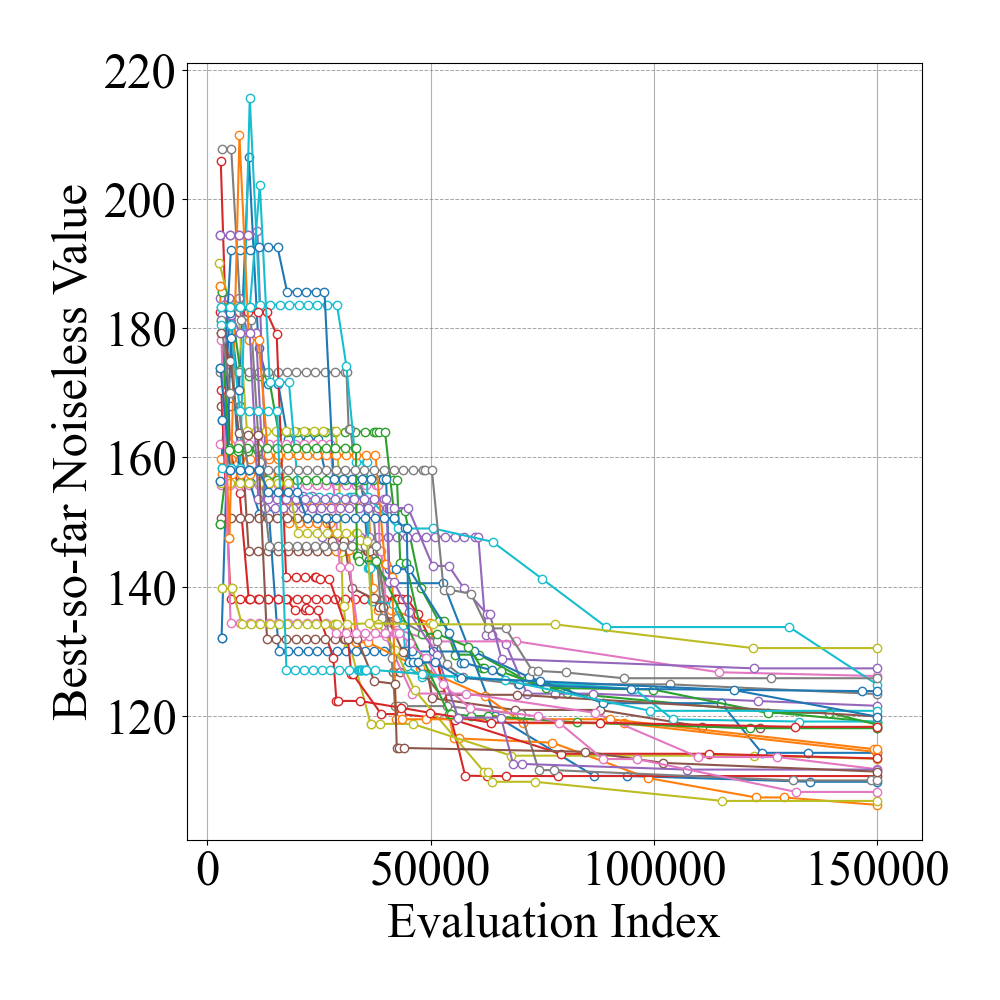}%
     \label{fig:true_mean_all_1}}
  \hfil
  \subfloat[Example 2]
    {\includegraphics[width=0.32\textwidth]{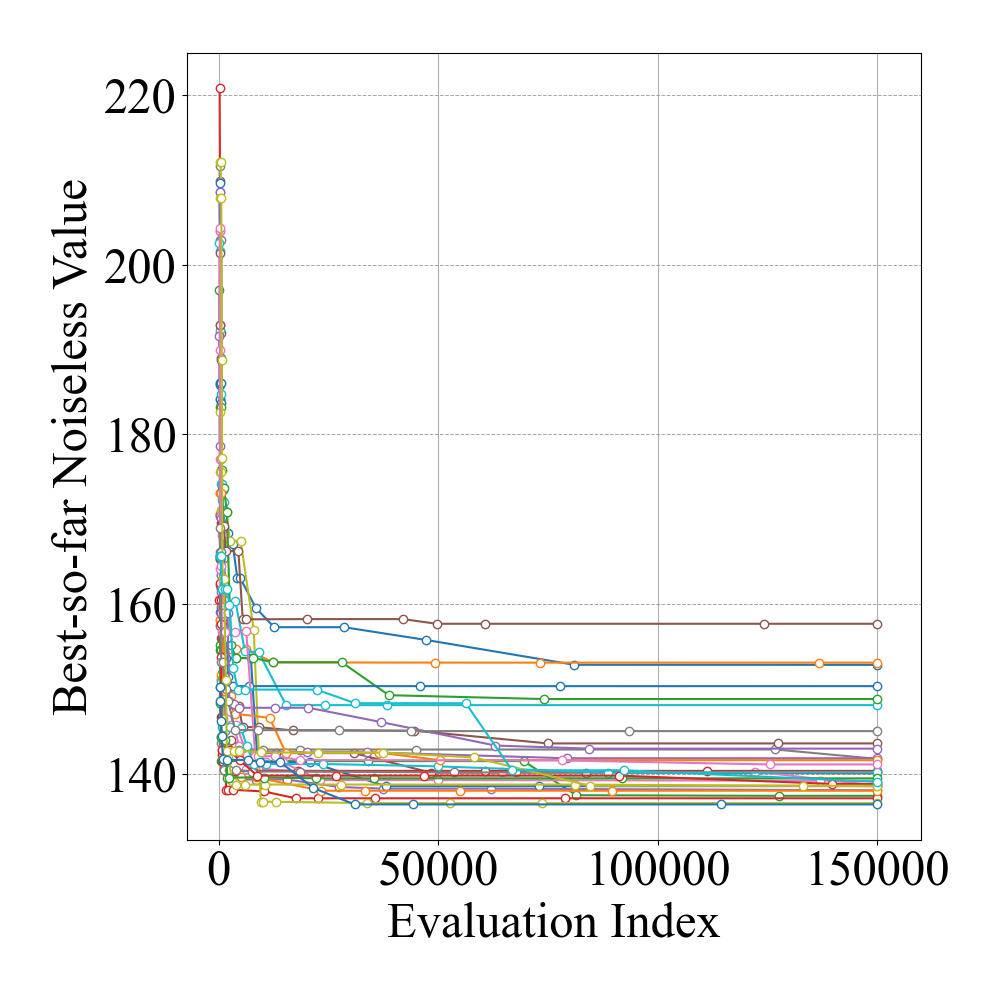}%
     \label{fig:true_mean_all_2}}
  \hfil
  \subfloat[Example 3]
    {\includegraphics[width=0.32\textwidth]{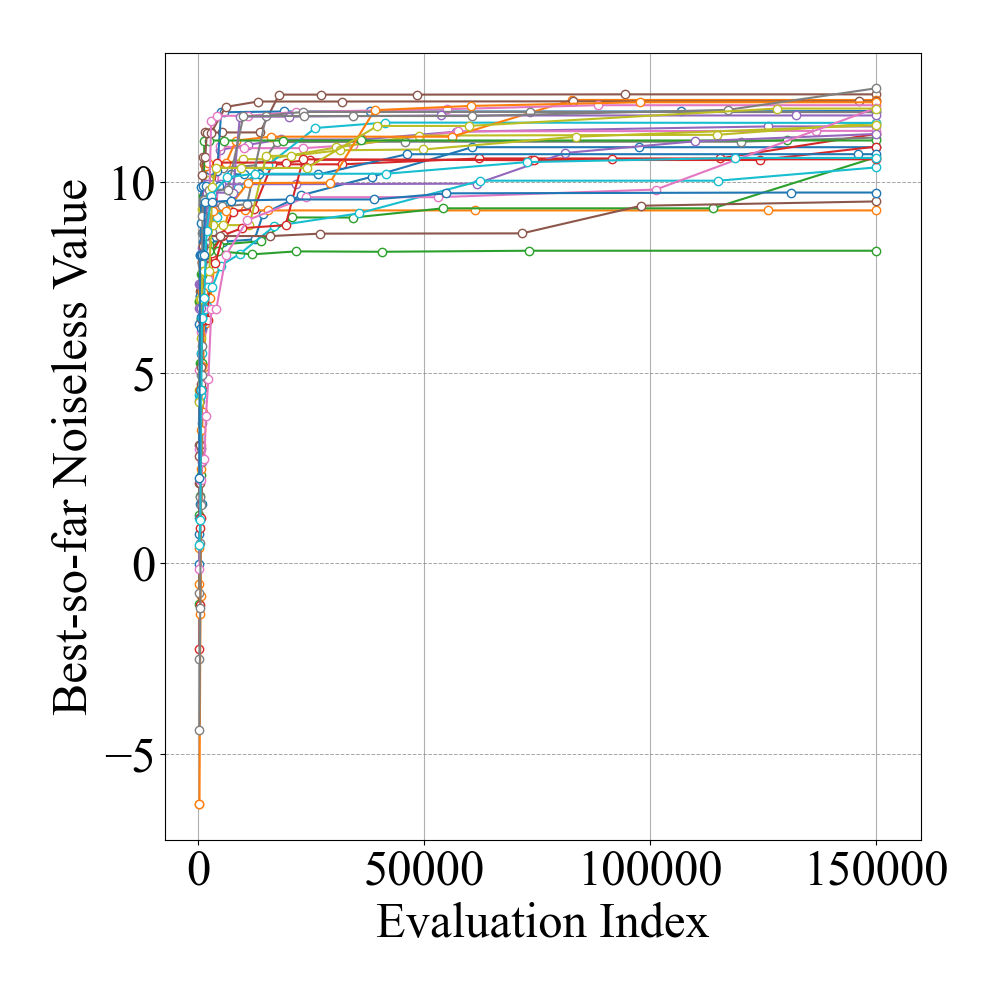}%
     \label{fig:true_mean_all_3}}
  \caption{CR-EA-C's first‐rank solution's noiseless value in each generation of all runs for Examples 1–3.}
  \label{fig:true_mean_all}
\end{figure*}

Figure~\ref{fig:convergence_median} presents the convergence curves corresponding to the median run of CR-EA-C and CR-EA-C-FD. Each plot reports the proposed best noiseless objective value $F(x)$ at every generation of the respective run. The results indicate that CR-EA-C clearly outperforms CR-EA-C-FD on all problems. For completeness, Figure~\ref{fig:true_mean_all} illustrates the convergence curves across all 31 runs of CR-EA-C, where each plot shows the proposed best noiseless $F(x)$ at every generation.

\begin{figure*}[!htbp]
  \centering
  \subfloat[Initialization]
    {\includegraphics[width=0.32\textwidth]{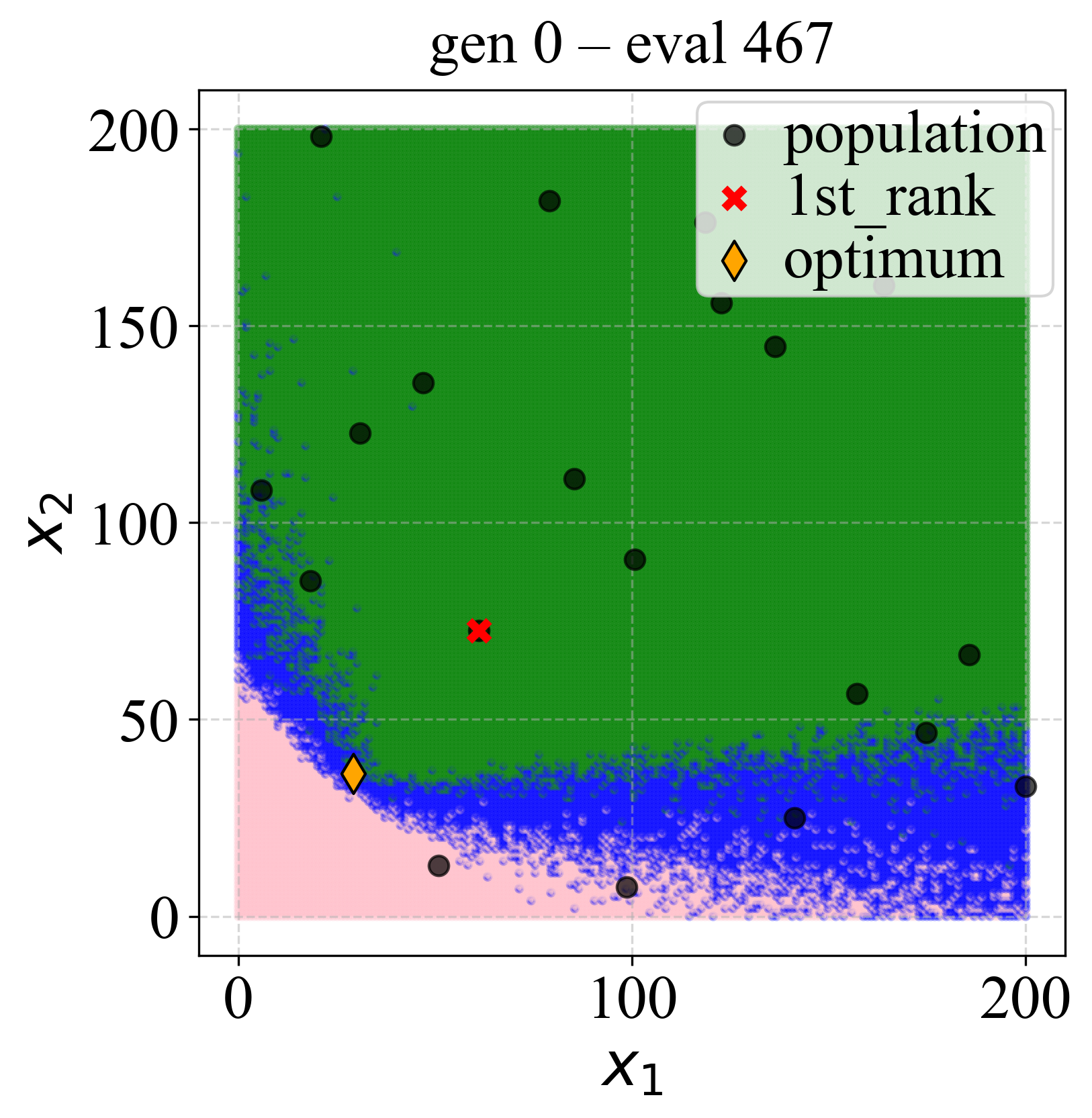}%
     \label{fig:ENRICO-FD movement 1}}
  \hfil
  \subfloat[Last generation of CR-EA-C]
    {\includegraphics[width=0.32\textwidth]{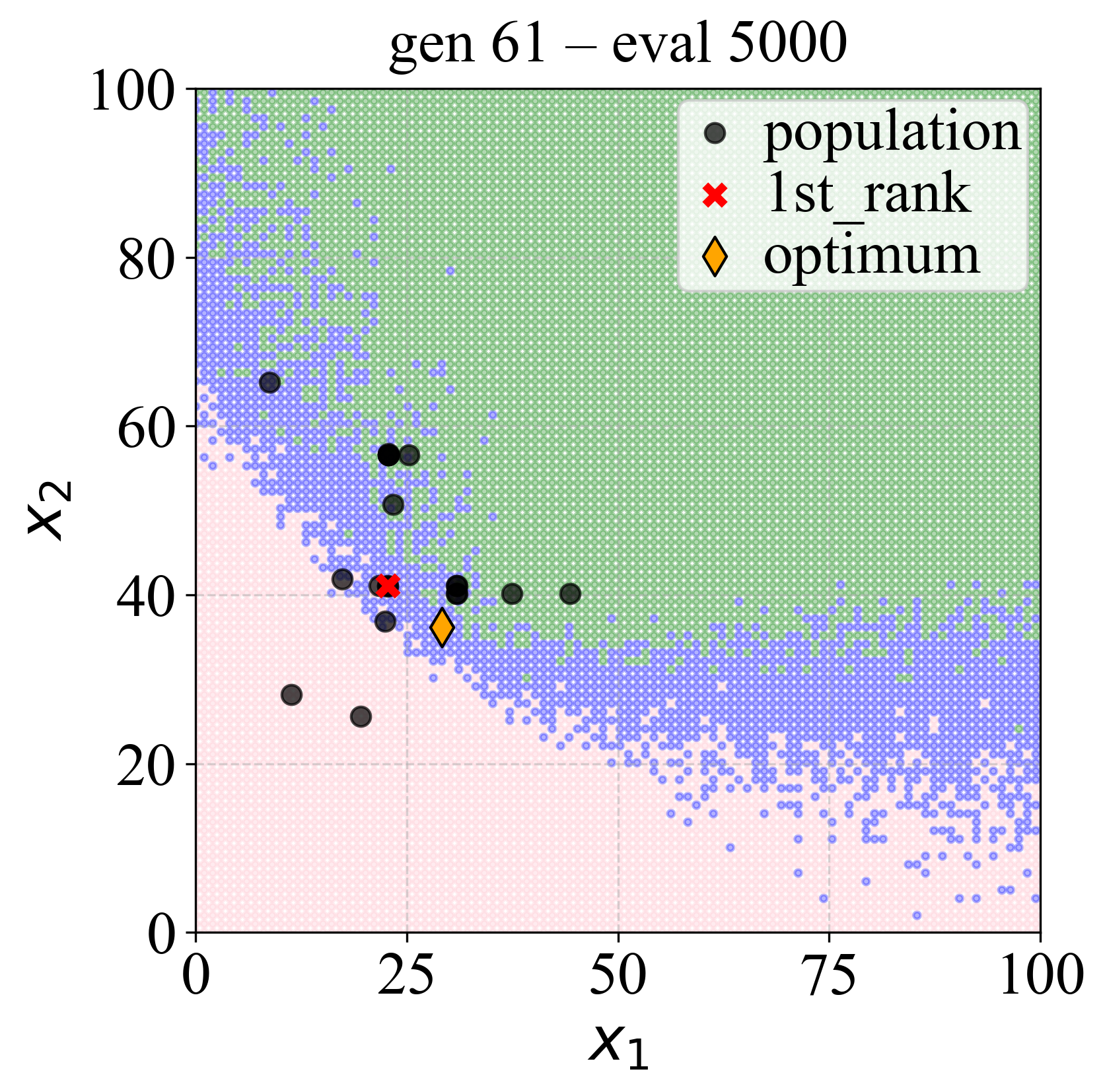}%
     \label{fig:ENRICO-FD movement 2}}
  \hfil
  \subfloat[Last generation of CR-EA-C-FD]
    {\includegraphics[width=0.32\textwidth]{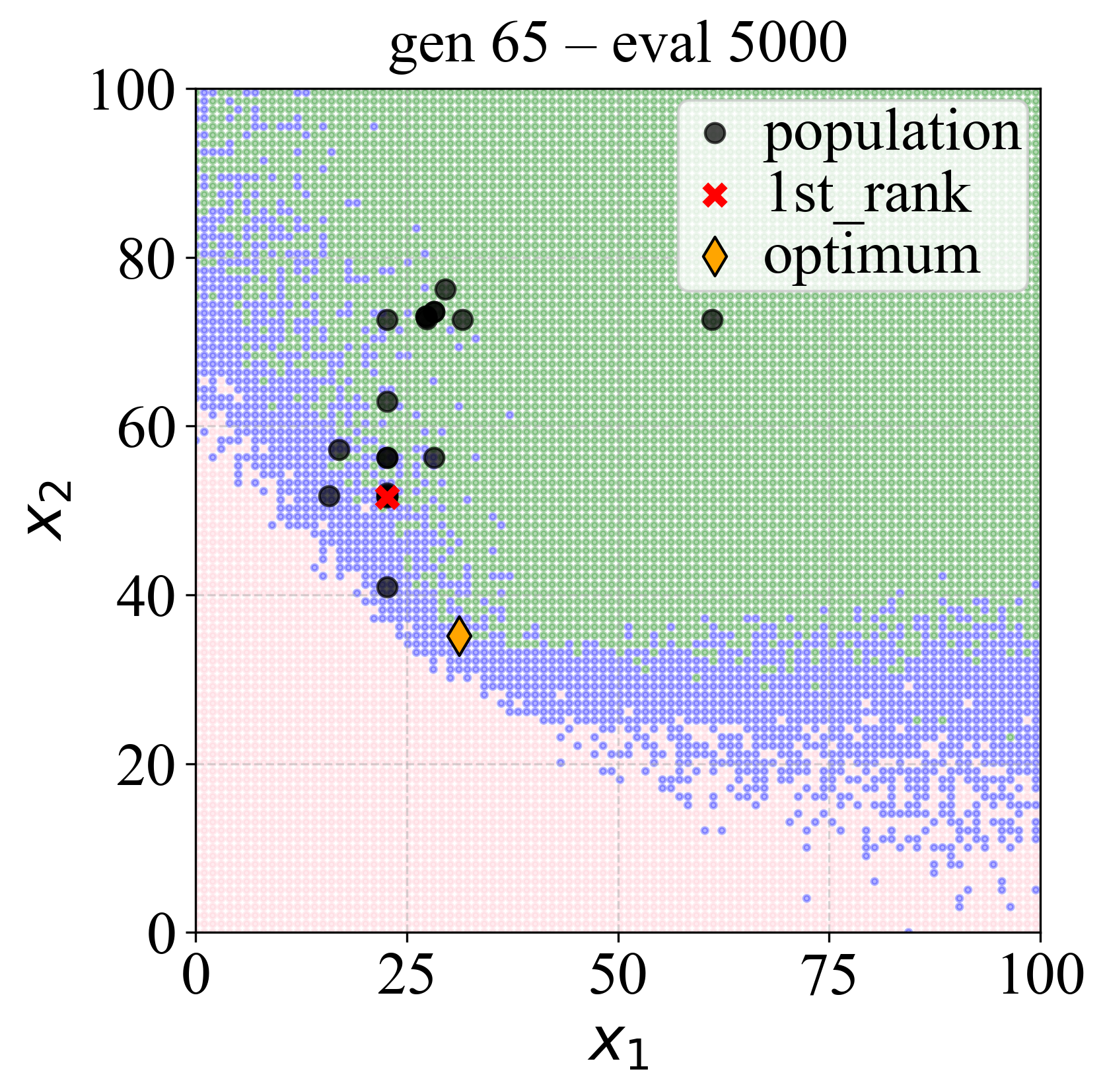}%
     \label{fig:ENRICO-FD movement 3}}
  \caption{The initial and last generation comparison with Examples 2.}
  \label{fig: ENRICO-FD movement}
\end{figure*}

To further highlight the behavioral differences between CR-EA-C and CR-EA-C-FD, Figure \ref{fig: ENRICO-FD movement} presents their search trajectories on a two-variable problem (Example~2) with bounds $[0, 200]^2$. In these figures, the feasible region is indicated in green, the \textit{maybe feasible} region in blue, and the infeasible region in red. All visualizations are based on 50 samples evaluated at each point.

Figure~\ref{fig:ENRICO-FD movement 1} shows the initial generation, where both CR-EA-C and CR-EA-C-FD start from identical positions. After performing the search with a total evaluation budget of 5000, the final generation of CR-EA-C (Fig.~\ref{fig:ENRICO-FD movement 2}) is observed to be closer to the optimum compared to CR-EA-C-FD (Fig.~\ref{fig:ENRICO-FD movement 3}). Notably, CR-EA-C maintains solutions both near the optimum and within infeasible regions, which contribute to guiding the population towards the optimum. This behavior is absent in CR-EA-C-FD (Fig.~\ref{fig:ENRICO-FD movement 3}).

\subsection{Parametric Study}
As previously mentioned in Section~\ref{sec: parameter settings}, the SBX crossover distribution index ($\eta_c$) of CR-EA-C is set to 30, whereas the original CR-EA~\cite{paper1} uses the value of 15. The rationale for increasing $\eta_c$ is to encourage smaller search steps, particularly when the population converges near the constraint boundary. This increases exploitation for better objective values and potentially reduces the impact of incorrectly classifying an infeasible solution as feasible when such errors occur. To investigate this hypothesis, four variants of CR-EA-C with $\eta_c$ values of 15, 20, 25, and 30 were evaluated on Examples 1, 2, and 3 using 31 independent runs and a budget of 150{,}000 function evaluations. 

Table~\ref{tab:parametric study} presents the results of this parametric study. To ensure a fair comparison, the best, worst, mean, standard deviation, Lower CI, and Upper CI values are computed using only the successful runs, with failed runs excluded from the statistics. The reported values are obtained by reevaluating each proposed final best solution of each run with 300{,}000 samples. Overall, these variants produce very similar results, indicating that CR-EA-C's performance is relatively insensitive to the choice of $\eta_c$. The parameter $\eta_c=30$ is selected as it offers a slightly better balance of objective values and average constraint violation (AV).

\begin{table*}[!htbp]
\centering
\caption{Parametric study results.}
\label{tab:parametric study}
\begin{tabular}{@{} ll c c c c c c c @{}}
\toprule
\makecell{\textbf{Problem}}
& \makecell{\textbf{$\eta_c$}}
& \textbf{Best} & \textbf{Worst} & \textbf{Mean} & \textbf{STDEV}
& \textbf{Lower CI} & \textbf{Upper CI} & \textbf{AV} \\
\midrule

\multirow{4}{*}{\makecell[c]{\textbf{Example 1}\\
\footnotesize\emph{Open Storage Network (minimize)}}}
& 15 & 107.6604 & 131.8255 & 118.2605 & 6.8941 & 115.9546 & 120.5669 & 0.0008 \\
& 20 & 110.1702 & 130.4887 & 119.3880 & 5.4513 & 117.6675 & 121.1079 & 0.0001 \\
& 25 & 108.0109 & 129.6900 & 118.8864 & 7.1020 & 116.5107 & 121.2619 & 0.0012 \\
& 30 & 105.6923 & 131.2129 & 116.7941 & 6.4725 & 114.6733 & 118.9167 & 0.0004 \\

\midrule

\multirow{4}{*}{\makecell[c]{\textbf{Example 2}\\
\footnotesize\emph{Oil Production Planning (minimize)}}}
& 15 & 136.4214 & 142.7890 & 138.3309 & 1.3243 & 137.9130 & 138.7490 & 0.0001 \\
& 20 & 136.9151 & 149.9942 & 139.5526 & 2.8642 & 138.6643 & 140.4389 & 0.0000 \\
& 25 & 136.4524 & 156.8308 & 139.2034 & 3.8822 & 137.9785 & 140.4282 & 0.0010 \\
& 30 & 136.7357 & 141.6885 & 138.4576 & 1.3909 & 138.0264 & 138.8885 & 0.0000 \\

\midrule

\multirow{4}{*}{\makecell[c]{\textbf{Example 3}\\
\footnotesize\emph{Multimodal Function Opt. (maximize)}}}
& 15 & -12.7618 & -10.7191 & -11.8449 & 0.5458 & -12.0144 & -11.6766 & 0.0000 \\
& 20 & -12.6240 & -10.3171 & -11.7025 & 0.5587 & -11.8763 & -11.5302 & 0.0000 \\
& 25 & -12.7627 & -10.0526 & -11.7606 & 0.7541 & -11.9941 & -11.5265 & 0.0000 \\
& 30 & -12.8448 &  -9.3194 & -11.7500 & 0.7342 & -11.9762 & -11.5218 & 0.0000 \\

\bottomrule
\end{tabular}
\end{table*}


\section{Additional Test Problems}
To further demonstrate the practicality of CR-EA-C in real-world scenarios, we evaluate it on two additional resource allocation problems. 

The first problem, which is taken from \cite{transporation_problem}, involves satisfying the demands of 10 customers, each with a random requirement, using goods supplied by 3 suppliers with fixed capacities. A cost matrix specifies the per-unit transportation cost for each supplier–customer pair. The objective is to minimize the total transportation cost while ensuring that all customer demands are met simultaneously with probability of at least $0.95$. The full mathematical formulation of this problem is provided in Appendix~\ref{appendix:Additional Problems}.

As baseline results~(and codes) from other peer methods used in previous section are not available for these problems, we compare CR-EA-C against a modified EA variants to assess the performance. The compared EA follows the same general structure as CR-EA-C: each solution is evaluated multiple times, and feasibility is determined using the Clopper–Pearson confidence interval. The key difference is that the EA uses a \emph{static} sampling size for all solutions and relies solely on the lower confidence bound to determine feasibility. The constraint violation (CV) of this EA is defined as
\begin{align}
CV(x) = \max\bigl(0, (1-\alpha) - \mathrm{LCB}(p(x))\bigr).
\label{eq:cv_for_EA}
\end{align}
Four EA variants are tested, each using a different fixed sampling size: 250, 500, 750, and 1000 samples per solution. These algorithm are denoted as EA-250, EA-500, EA-750, and EA-1000 respectively. The population size and all reproduction operator settings are kept identical to CR-EA-C. All algorithms are independently run 31 times with population size of 100. Two evaluation budgets are considered for individual runs: 2,500,000 and 4,000,000.

\begin{figure}[!ht]
    \centering
    \includegraphics[width=0.8\linewidth]{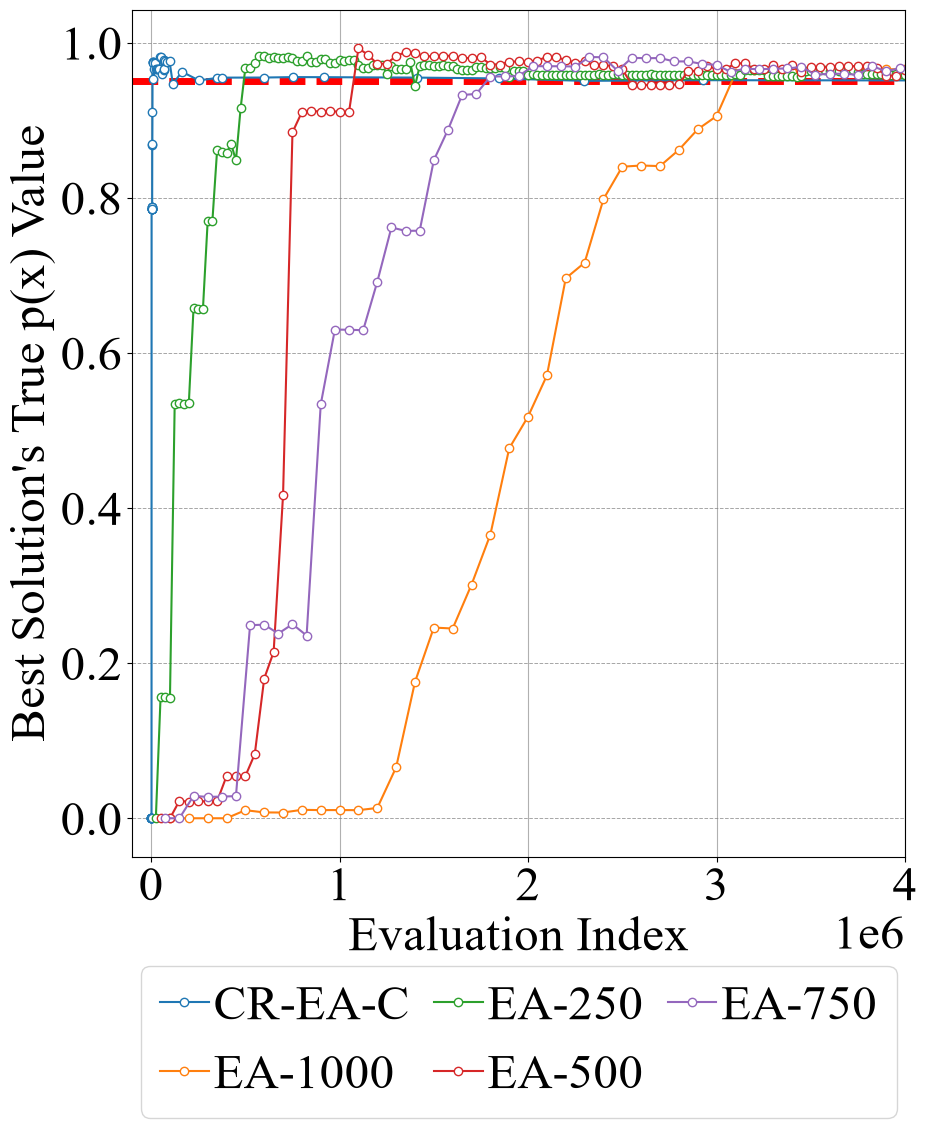}
    \caption{Median run's first‐rank solution's $p(x)$ value in each generation for Transportation problem~(computed using large sample size of 300,000 evaluations). Dashed red line is at $p(x)=0.95$}
    \label{fig:convergence_feasibility}
\end{figure}

Figure~\ref{fig:convergence_feasibility} shows the convergence of the first-rank solution’s true feasibility probability $p(x)$ for the five algorithms on the first problem. The representative value of true $p(x)$ is obtained by reevaluating the solution 300,000 times. From the plot, it is evident that CR-EA-C identifies a feasible solution substantially faster than all four EA variants, even when compared to the EA-250 which has the fastest convergence out of all 4 variants. Notably, CR-EA-C achieves feasibility using well under 500{,}000 evaluations.

Table~\ref{tab:results_table_transport} provides a performance summary similar to Table~\ref{tab:results_table}, with the addition of \#fails column, which reports the number of runs that fail to find a feasible solution within the evaluation budget. Across both evaluation budgets considered, CR-EA-C achieves a perfect success rate, with all 31 runs producing feasible solutions. In contrast, every modified EA variant exhibits multiple failed runs, even under the larger budget.

The best, worst, mean, standard deviation, Lower CI, and Upper CI values are computed using only the successful runs for all algorithms, with failed runs excluded from the statistics. Notably, EA variants continue to improve as the evaluation budget increases, while CR-EA-C appears to plateau around 2{,}500{,}000 evaluations with no further improvement observed at 4{,}000{,}000. However, even under this stricter evaluation, CR-EA-C maintains competitive objective performance. Overall, these results highlight CR-EA-C's capability of delivering reliable feasibility while preserving competitive solution's objective value, illustrating a good trade-off between robustness and optimality.

The second problem is taken from \cite{poojari} and concerns a pension fund allocation setting. A company must meet its liabilities over the next 15 years, where the liability amount in each year follows a random distribution. These liabilities are to be covered by investing an initial capital of 250,000 in 3 different types of bonds. Each bond has different price to buy and will have different yield every year. The objective is to maximize the total return on investment while ensuring that all yearly payments are satisfied with a joint probability of at least $0.95$. The full mathematical formulation of this problem is provided in Appendix~\ref{appendix:Additional Problems}. Based on~\cite{poojari}, the known optimal solution is reported as $\{65.8, 83.7, 86.2\}$ with an objective value of $98{,}160$. CR-EA-C, EA-250, EA-500, EA-750, and EA-1000 are evaluated on this problem using 31 independent runs, each with evaluation budgets of $100{,}000$ and $150{,}000$, and a population size of 30.

The results are summarised in Table~\ref{tab:pension_fund_results} and were obtained by reevaluating the proposed best solution of each run 300,000 times. For consistency with Table~\ref{tab:results_table_transport}, failed runs are also excluded from all the statistics. CR-EA-C records one failed run under both evaluation budgets. However, it outperforms all EA variants across all other reported metrics for both budgets. Moreover, the best run of CR-EA-C in each budget achieves an objective value close to the reported global optimum solution. These findings showcase CR-EA-C’s ability to deliver stronger objective-value performance than the EA variants, albeit with a minor chance of failed runs~(which is not unique to CR-EA-C).

\begin{table*}[!htbp]
\centering
\caption{Performance summary of algorithms tested on transportation problem.}
\label{tab:results_table_transport}
\begin{tabular}{@{} ll c c c c c c c c @{}}
\toprule
\textbf{Problem}
& \textbf{Algorithm}
& \textbf{Best} & \textbf{Worst} & \textbf{Mean} & \textbf{STDEV}
& \textbf{Lower CI} & \textbf{Upper CI} & \textbf{AV} & \textbf{\#fails} \\
\midrule

\multirow{5}{*}{\makecell[c]{
\textbf{transportation}\\
\footnotesize\emph{(evaluation budget = 2500000)}\\
\footnotesize\emph{(minimize)}
}}
  & CR-EA-C
    & 56409.62
    & 71017.70
    & 63508.19
    & 3640.57
    & 62380.06
    & 64636.31
    & 0.0000
    & 0 \\

  & EA-250
    & 49142.03
    & 68493.86
    & 57261.96
    & 4299.62
    & 55823.75
    & 58700.18
    & 0.1226
    & 4 \\

  & EA-500
    & 48678.31
    & 66416.83
    & 59479.20
    & 4291.67
    & 58043.64
    & 60914.76
    & 0.1222
    & 4 \\

  & EA-750
    & 55054.56
    & 74181.91
    & 63698.17
    & 4697.11
    & 62093.50
    & 65302.83
    & 0.1071
    & 5 \\

  & EA-1000
    & 57568.37
    & 74710.43
    & 64243.87
    & 4717.67
    & 62253.40
    & 66234.33
    & 0.0815
    & 13 \\

\midrule

\multirow{5}{*}{\makecell[c]{
\textbf{transportation}\\
\footnotesize\emph{(evaluation budget = 4000000)}\\
\footnotesize\emph{(minimize)}
}}
  & CR-EA-C
    & 56340.59
    & 71017.70
    & 63395.60
    & 3656.82
    & 62262.44
    & 64528.76
    & 0.0000
    & 0 \\

  & EA-250
    & 47856.94
    & 67429.24
    & 56270.94
    & 4309.79
    & 54765.82
    & 57776.05
    & 0.1229
    & 6 \\

  & EA-500
    & 46236.00
    & 71767.76
    & 57462.07
    & 4957.85
    & 55836.90
    & 59087.25
    & 0.0918
    & 3 \\

  & EA-750
    & 52744.30
    & 69738.62
    & 59518.36
    & 4133.35
    & 58163.45
    & 60873.27
    & 0.0919
    & 3 \\

  & EA-1000
    & 52612.33
    & 69277.37
    & 60564.09
    & 4430.63
    & 59111.74
    & 62016.44
    & 0.0328
    & 3 \\

\bottomrule
\end{tabular}
\end{table*}

\begin{table*}[!htbp]
\centering
\caption{Performance summary of algorithms tested on pension\_fund problem.}
\label{tab:pension_fund_results}
\begin{tabular}{@{} ll c c c c c c c c @{}}
\toprule
\textbf{Problem}
& \textbf{Algorithm}
& \textbf{Best} & \textbf{Worst} & \textbf{Mean} & \textbf{STDEV}
& \textbf{Lower CI} & \textbf{Upper CI} & \textbf{AV} & \textbf{\#fails} \\
\midrule

\multirow{5}{*}{\makecell[c]{
\textbf{pension\_fund}\\
\footnotesize\emph{(evaluation budget = 100000)}\\
\footnotesize\emph{(maximize)}
}}
  & CR-EA-C
    & 97944.63
    & 87390.05
    & 93013.61
    & 3231.60
    & 91993.98
    & 94033.25
    & 0.0003
    & 1 \\

  & EA-250
    & 93295.27
    & 55347.91
    & 85085.94
    & 7586.88
    & 82734.94
    & 87436.93
    & 0.0000
    & 0 \\

  & EA-500
    & 93385.53
    & 35947.28
    & 72392.51
    & 14230.63
    & 67982.78
    & 76802.24
    & 0.0000
    & 0 \\

  & EA-750
    & 90272.62
    & 29896.59
    & 64109.42
    & 16145.41
    & 58708.80
    & 69510.05
    & 0.0212
    & 4 \\

  & EA-1000
    & 95989.87
    & 35045.03
    & 66389.58
    & 16738.57
    & 60671.21
    & 72107.95
    & 0.0188
    & 5 \\

\midrule

\multirow{5}{*}{\makecell[c]{
\textbf{pension\_fund}\\
\footnotesize\emph{(evaluation budget = 150000)}\\
\footnotesize\emph{(maximize)}
}}
  & CR-EA-C
    & 97944.63
    & 87390.05
    & 93105.57
    & 3125.51
    & 92119.41
    & 94091.73
    & 0.0003
    & 1 \\

  & EA-250
    & 93805.07
    & 68936.06
    & 88031.30
    & 5526.63
    & 86318.74
    & 89743.87
    & 0.0000
    & 0 \\

  & EA-500
    & 93640.53
    & 42794.82
    & 81388.24
    & 10798.27
    & 78042.12
    & 84734.37
    & 0.0000
    & 0 \\

  & EA-750
    & 91194.78
    & 42163.76
    & 72656.88
    & 14093.54
    & 68210.09
    & 77103.67
    & 0.0066
    & 1 \\

  & EA-1000
    & 89298.24
    & 47938.90
    & 75300.01
    & 11607.18
    & 71568.50
    & 79031.53
    & 0.0007
    & 2 \\

\bottomrule
\end{tabular}
\end{table*}


\section{Conclusion}
In this paper, we presented CR-EA-C, an evolutionary algorithm designed for noisy black-box optimization under joint chance constraints. CR-EA-C integrates several key innovations: analytical approximations for constraint handling, a novel ranking and selection mechanism, an OCBA-driven resampling strategy, and a modified infeasibility-driven survival strategy. 

Extensive experiments on three benchmark problems (one mathematical test problem and two practical applications problem), together with two additional real-world resource allocation problems, demonstrate that CR-EA-C is able to consistently satisfy joint chance constraints while achieving competitive or superior objective values when compared to other algorithms. These results confirm the efficacy of our approach and highlight its reliability as a general-purpose solver for black-box optimization problems under uncertainty.

For future work, we plan to extend the resampling strategy and uncertainty-handling mechanism to Bayesian optimization framework, with the aim of substantially reducing the number of function evaluations and enhancing practical applicability for computationally expensive problems. Additionally, we encourage the integration of our modular procedures into other population-based metaheuristics, opening opportunities for broader improvements in noisy constrained optimization.

\section*{Acknowledgment} 
EH gratefully acknowledges the University of New South Wales for its support through the Tuition Fee Scholarship (TFS), which made this research possible. HS and TR acknowledge support from Australian Research Council (ARC) Discovery Project DP220101649. Generative AI tools were used to assist with improving the clarity, readability, and expression of the writing of this work.


\nocite{*} 

\bibliographystyle{asmejour}   

\bibliography{refs} 





\appendix   

\section{Benchmark Problems}
\label{appendix:Benchmark Problems}

\textbf{Example 1} (Open storage networks) \cite{example_1_2}
\begin{align*}
\min_{x,\eta}\;&\mathbb{E}\bigl[\eta + \sum_{i=1}^9 |x_i|\bigr] \\
\text{s.t.}\\
\;&\Pr\!\left\{
  \begin{aligned}
    70 - \sum_{i=1}^4 x_i + \xi_1 &\ge 10,\\
    70 - \sum_{i=1}^4 x_i + \xi_1 &\le 120,\\
    80 + x_2 - x_5 - x_6 + \xi_2 &\ge 20,\\
    80 + x_2 - x_5 - x_6 + \xi_2 &\le 100,\\
    60 + x_3 - x_7 - x_8 + \xi_3 &\ge 10,\\
    60 + x_3 - x_7 - x_8 + \xi_3 &\le 80,\\
    50 + x_4 + x_6 + x_8 - x_9 + \xi_4 &\ge 0,\\
    50 + x_4 + x_6 + x_8 - x_9 + \xi_4 &\le 90
  \end{aligned}
\!\right\}
\;\ge\;0.9,\\[1ex]    
&10 \le x_{1} \le 50,\quad
0 \le x_{2} \le 10,\quad
0 \le x_{3} \le 10,\\
&0 \le x_{4} \le 15,\quad
15 \le x_{5} \le 60,\quad
-5 \le x_{6} \le 5,\\
&15 \le x_{7} \le 60,\quad
-5 \le x_{8} \le 5,\quad
20 \le x_{9} \le 70,\\[1ex]
&\xi_{1}\sim \log\mathcal{N}(2.24,1.12),\quad
 \xi_{2}\sim \log\mathcal{N}(1.60,1.28),\\
&\xi_{3}\sim \log\mathcal{N}(1.87,1.45),\quad
 \xi_{4}\sim \log\mathcal{N}(1.30,1.34),\\
&\eta \sim \mathcal{N}(0,2).
\end{align*}

\vspace{1em} 

\textbf{Example 2} (Oil production planning) \cite{example_1_2}
\begin{align*}
  \min_{x,\eta}\;&\mathbb{E}\bigl[\eta + 2x_1 + 3x_2\bigr] \\
  \text{s.t.}\\
  &\Pr\!\left\{
    \begin{aligned}
      (2 + \xi_1)\,x_1 \;+\; 6\,x_2 &\ge 180 + \xi_3,\\
      3\,x_1 \;+\; (3.4 - \xi_2)\,x_2 &\ge 162 + \xi_4
    \end{aligned}
  \!\right\}
  \;\ge\;0.8,\\[1ex]
  &x_1 + x_2 \le 100,\quad x_1 \ge 0,\quad x_2 \ge 0,\\
  &\xi_{1}\sim U(-0.8,0.8),\quad
   \xi_{2}\sim \exp(0.4),\\
  &\xi_{3}\sim N(0,12),\quad
   \xi_{4}\sim N(0,9),\quad
   \eta \sim N(0,2).
\end{align*}

\vspace{1em} 

\textbf{Example 3} (Multimodal function optimization) \cite{example_3}
\begin{align*}
  \max_{x,\eta}\;&\mathbb{E}\bigl[\eta + \sum_{k=1}^3 x_k\sin(k\pi x_k)\bigr] \\
  \text{s.t.}\\
  &\Pr\!\left\{
    \begin{aligned}
      \xi_1\,x_1 \;+\;\xi_2\,x_2 \;+\;\xi_3\,x_3 \;-\;10 &\le 0,\\
      \xi_1\,x_1^2 + \xi_2\,x_2^2 + \xi_3\,x_3^2 \;-\;100 &\le 0
    \end{aligned}
  \!\right\}
  \;\ge\;0.7,\\[1ex]
  &\xi_{1}\sim U(0.8,1.2),\quad
   \xi_{2}\sim U(1,1.3),\\
  &\xi_{3}\sim U(0.8,1.0), \quad
  \sigma_{1}\sim N(1,0.5),\\
  &\sigma_{2}\sim \exp(1.2),\quad
  \sigma_{3}\sim \log\mathcal{N}(0.8,0.6),\\
   &\eta \sim N(0,2).
\end{align*}

\section{Additional Problems}
\label{appendix:Additional Problems}

\textbf{Transportation Problem} \cite{transporation_problem}
\begin{align*}
\min_{x}\;& \sum_{i=1}^{3}\sum_{j=1}^{10} c_{ij}\,x_{ij} \\
\text{s.t.}\quad
&\Pr\!\left\{
\begin{aligned}
\sum_{i=1}^{3} x_{ij} \;\ge\; b_j,\quad j = 1,\ldots,10
\end{aligned}
\right\}
\;\ge\; 0.95,\\[1ex]
&\sum_{j=1}^{10} x_{ij} \;\le\; M_i,\quad i = 1,2,3,\\
&x_{ij} \ge 0,\quad i=1,2,3,\; j=1,\ldots,10,\\
&b_j \sim \mathcal{N}(1000,100),\quad j=1,\ldots,10.
\end{align*}
\noindent
with parameter values of
\begin{align*}
&M_i = 5000,\quad i = 1,2,3,\\
&p = 0.95.
\end{align*}
\[
\resizebox{\columnwidth}{!}{$
C =
\begin{pmatrix}
4.37 & 9.56 & 7.59 & 6.39 & 2.40 & 2.40 & 1.52 & 8.80 & 6.41 & 7.08 \\
1.19 & 9.73 & 8.49 & 2.91 & 2.64 & 2.65 & 3.74 & 5.72 & 4.89 & 3.62 \\
6.51 & 2.26 & 3.63 & 4.30 & 5.10 & 7.93 & 2.80 & 5.63 & 6.33 & 1.42
\end{pmatrix}
$}
\]

\noindent
where
\begin{align*}
&x_{ij}
&&\text{amount shipped from supplier $i$ to customer $j$} \\
&c_{ij}
&&\text{unit transportation cost from supplier $i$ to customer $j$} \\
&M_i
&&\text{capacity of supplier $i$} \\
&b_j 
&&\text{random demand of customer $j$}
\end{align*}

\vspace{1em} 

\textbf{Pension Fund Problem} \cite{poojari}
\begin{align*}
\max_{x}\;& \sum_{i=1}^{P} a_{iT}\,x_i \\
\text{s.t.}\quad
&\Pr\!\left\{
\begin{aligned}
\sum_{i=1}^{P} a_{ij}\,x_i \;\ge\; b_j,\quad j = 1,\ldots,T
\end{aligned}
\right\}
\;\ge\; 0.95,\\[1ex]
&x_i \ge 0,\quad i = 1,\ldots,P,\\[1ex]
&a_{ij} = \sum_{k=1}^{j} \alpha_{ik} - \gamma_i,\quad i=1,\ldots,P,\; j=1,\ldots,T,\\
&b_j = \sum_{k=1}^{j} \beta_k - B,\quad j=1,\ldots,T.
\end{align*}
\noindent
with parameter values of
\begin{align*}
&T = 15, \\
&P = 3, \\
&B = 250{,}000, \\
&\gamma = \{980,\,970,\,1050\},\\
&\beta = \{11{,}000,\,12{,}000,\,14{,}000,\,15{,}000,\,16{,}000,\,18{,}000,\\
&\qquad\;\;20{,}000,\,21{,}000,\,22{,}000,\,24{,}000,\,25{,}000,\,30{,}000,\\
&\qquad\;\;31{,}000,\,31{,}000,\,31{,}000\},\\
&\alpha =
\begin{pmatrix}
0   & 0   & 0   \\
60  & 65  & 75  \\
60  & 65  & 75  \\
60  & 65  & 75  \\
60  & 65  & 75  \\
1060 & 65 & 75  \\
0   & 65  & 75  \\
0   & 65  & 75  \\
0   & 65  & 75  \\
0   & 65  & 75  \\
0   & 65  & 75  \\
0   & 1060 & 75 \\
0   & 0   & 75  \\
0   & 0   & 75  \\
0   & 0   & 1075
\end{pmatrix}.
\end{align*}
\noindent
where
{\setlength{\jot}{2pt}
\begin{align*}
&T 
&&\text{number of time periods (years)} \\
&P 
&&\text{number of bond types} \\
&B 
&&\text{initial capital} \\
&\gamma_i 
&&\text{cost of purchasing one unit of bond $i$} \\
&\beta_j 
&&\text{liability payment in year $j$} \\
&\alpha_{ij}
&&\text{yield of bond $i$ in year $j$} \\
&a_{ij}
&&\text{cumulative net return of bond $i$ up to year $j$} \\
&b_j
&&\text{cumulative liability shortfall up to year $j$}
\end{align*}
}


\end{document}